\documentclass[journal]{IEEEtran}
\usepackage{amsmath,amsfonts}
\usepackage{amsthm}
\usepackage{algorithmic}
\usepackage{algorithm}
\usepackage{array}
\usepackage[caption=false,font=normalsize,labelfont=sf,textfont=sf]{subfig}
\usepackage{textcomp}
\usepackage{stfloats}
\usepackage{url}
\usepackage{verbatim}
\usepackage{graphicx}
\usepackage{cite}
\usepackage{booktabs}
\usepackage{multirow}
\usepackage{graphicx}

\usepackage{makecell}

\usepackage{bm}
\usepackage{hyperref}
\usepackage[table]{xcolor}
\usepackage{tabularx}

\theoremstyle{remark}

\makeatletter
\def\th@remark{%
  \normalfont % 正文是直立体
  \thm@headfont{\bfseries\itshape}% 标题加粗+斜体
}
\makeatother

\theoremstyle{definition}

\theoremstyle{definition}

\theoremstyle{plain}

\title{\LARGE \bf
ContactWorld: What Representations Matter for Vision--Tactile Latent World Models in Contact-Rich Manipulation
}

\author{Author Names Omitted}

\author{
Zhiyuan Zhang$^{1,*}$,
Pokuang Zhou$^{1,*}$,
Kaidi Zhang$^{1}$,
Adeesh Desai$^{1}$,
Temitope Amosa$^{1}$, \\
Davood Soleymanzadeh$^{2}$,
Jiuzhou Lei$^{2}$,
Yuhao Zhou$^{1}$,
Minghui Zheng$^{2}$,
Yu She$^{1,\dagger}$ \\[0.8em]
\thanks{This work was partially supported by the United States Department of Agriculture (USDA) under Grant Nos. 2023-67021-39072 and 2024-67021-42878, and by the National Science Foundation (NSF) under Grant Nos. 2423068 and 2520136. The authors also thank Daimon Robotics for providing the DM-Tac W2M tactile sensor used in the real-world experiments.}
\thanks{$^{*}$These authors contributed equally. 
$^{\dagger}$Corresponding author.}%
\thanks{Zhiyuan, Pokuang, Kaidi, Adeesh, Temitope, Yuhao, and Yu are with the Department of Industrial Engineering, Purdue University, West Lafayette, IN 47907, USA 
{\tt\footnotesize \{zhan5570, zhou1458, zhan5896, desai274, tamosa, zhou1437, shey\}@purdue.edu}}%
\thanks{Davood, Jiuzhou, and Minghui are with the Department of Mechanical Engineering, Texas A\&M University, College Station, TX 77843, USA 
{\tt\footnotesize \{davoodso, jiuzl, mhzheng\}@tamu.edu}}%
}

\begin{document}

\maketitle

\begin{abstract}
Contact-rich manipulation poses a fundamental challenge for world models: visual and tactile observations capture different aspects of physical interaction, and their utility depends critically on how this information is represented.
We introduce \textbf{ContactWorld}, a controlled benchmark and empirical study of vision--tactile representations across 12 contact-rich manipulation tasks.
Using a fixed world-model architecture, training procedure, and planning framework, we examine representation effects through three complementary properties: \textit{spatial fidelity}, \textit{motion coherence}, and \textit{predictive stability}.
Point clouds preserve task-relevant geometry and track physical motion more reliably than image observations, helping explain their higher average planning success of 32.1\%, compared with 20.7\% and 22.0\% for wrist- and front-view RGB, respectively.
Tactile observations generally reduce long-horizon prediction-error accumulation, but these gains do not translate uniformly into task success.
Structured tactile force fields provide the most consistent downstream improvements, with PointCloud+TacFF achieving the highest overall simulated success rate of 36.1\%.
We further validate these trends through 900 physical trials spanning six tasks, two robot platforms, and three tactile sensing systems.
Together, ContactWorld identifies sensory representation as a central design factor in vision--tactile world models and provides empirical guidance for predictive planning in contact-rich manipulation.
\vspace{0.5em}
\par
\noindent Project website: \url{https://contact-world.github.io}
\end{abstract}


\section{Introduction}

\begin{figure}[t]
    \centering
    \includegraphics[width=\columnwidth]{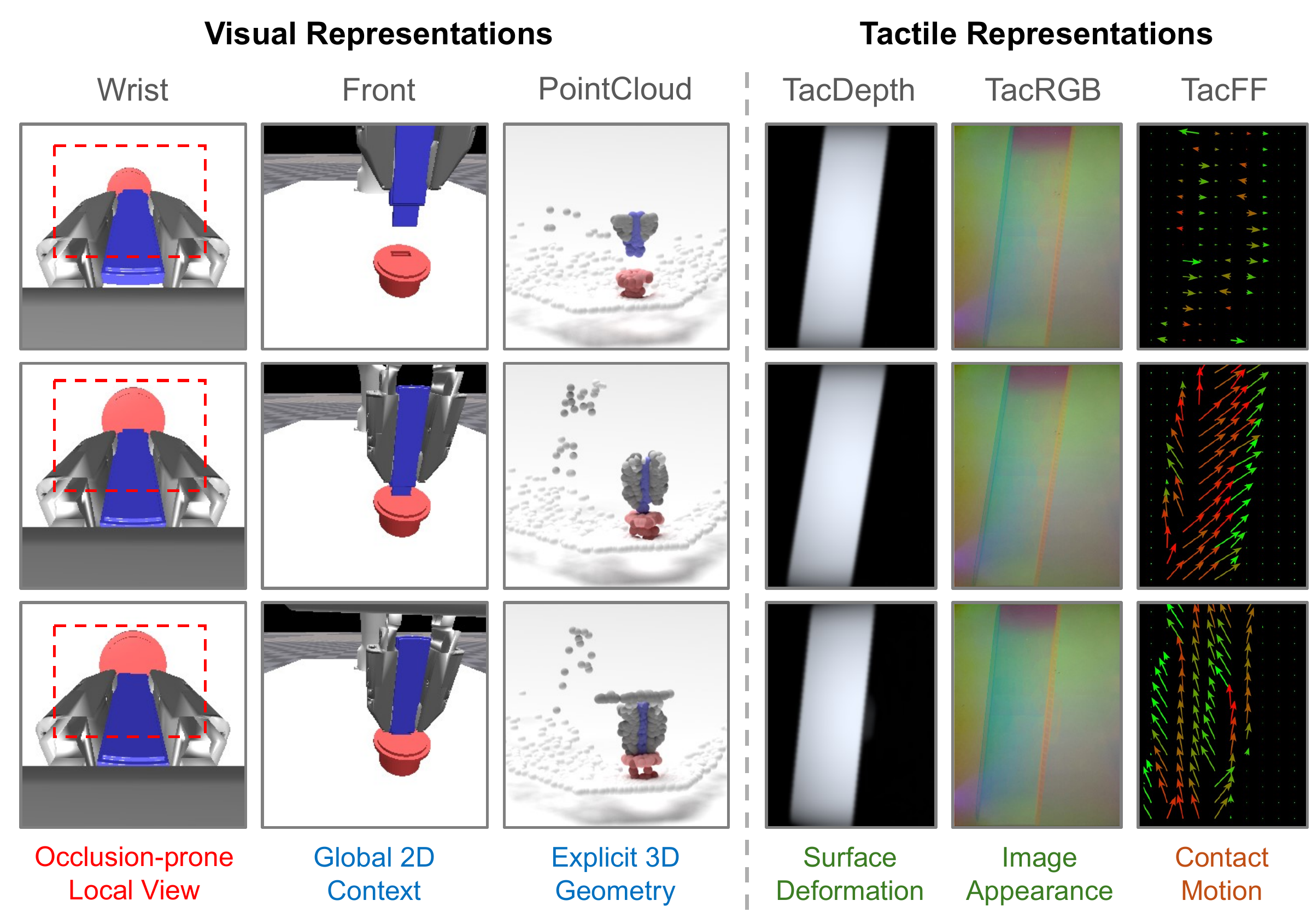}
    \caption{\textbf{Visual and tactile representations during contact-rich
    interaction.}
    Representative observations at three stages of insertion illustrate
    differences in spatial structure and contact dynamics across modalities.}
    \label{fig:modality_vis}
\end{figure}

Contact-rich manipulation remains a central challenge in robot learning because
successful interaction requires reasoning about constrained geometry, local
contact events, friction, partial observability, and temporally evolving
multimodal feedback~\cite{luo2025tactile,pattabiraman2024learning,
xue2025reactive,zhao2025touch}.
Unlike predominantly vision-driven manipulation, contact-rich interaction
tightly couples perception and action through changing physical constraints,
such that small state-estimation or prediction errors can lead to large
execution failures.
Vision provides global scene context but is vulnerable to occlusion and
geometric ambiguity, whereas touch provides local measurements of contact and
deformation through physical interaction.
As illustrated in Fig.~\ref{fig:modality_vis}, representations of the same
interaction can therefore exhibit substantially different spatial and temporal
structures.

Recent latent world models support robotic planning by learning compact,
action-conditioned dynamics for future prediction and latent-space
optimization~\cite{assran2025v,maes2026leworldmodel,terver2025drives,
luo2026being,zheng2025flare,jha2026reconstruction,dinowm}.
Joint-embedding objectives are particularly attractive because they learn
predictive representations without reconstructing raw observations.
Meanwhile, modern tactile sensors capture local geometry, deformation, and
force that may be difficult to infer from vision
alone~\cite{luo2025tactile,higuera2026visuo,gelsight,
luu2025manifeel,huang20243d}.
These developments create an opportunity for vision--tactile world models, but
also raise a fundamental question: \emph{which sensory representations support
reliable prediction and planning in contact-rich interaction?}

Existing work primarily emphasizes model architecture, policy objectives,
sensing hardware, or aggregate task performance
~\cite{dinowm,gao2025flip,higuera2026visuo,
zheng2026omnivta,lou2026dream}.
Sensory representation is often treated as an implementation choice, leaving
unclear why the same predictive framework produces different planning outcomes
across observation modalities.
This issue is especially important in contact-rich manipulation, where a useful
representation must preserve task-relevant geometry, respond meaningfully to
physical interaction, and remain predictable over extended horizons.
Likewise, the benefit of touch depends not only on its availability, but also
on how contact information is represented and how it complements vision.

We introduce \textbf{ContactWorld}, a controlled benchmark and empirical study
of sensory representation in vision--tactile world models.
It contains 12 task variants spanning insertion, disassembly, screwing, and
exploratory interaction, and evaluates wrist-view RGB, front-view RGB, and
point clouds together with tactile RGB, tactile depth, and tactile force
fields.
All configurations use the same latent world-model architecture, training
procedure, receding-horizon planner, and evaluation protocol, allowing their
differences to be attributed primarily to observation representation.

Beyond task success, we analyze representations through three complementary
properties.
\emph{Spatial fidelity} measures whether latent features preserve
task-relevant spatial relations;
\emph{motion coherence} measures whether representational changes correspond
to physical scene motion; and
\emph{predictive stability} measures error accumulation over autoregressive
rollouts.
Together, these properties connect representation structure to learned
dynamics and downstream planning performance.

The results show that point clouds preserve task-relevant geometry, align
latent changes with physical motion, and maintain stable predictions more
reliably than image observations.
Their average planning success reaches 32.1\%, compared with 20.7\% and 22.0\%
for wrist- and front-view RGB, respectively, with the advantage increasing at
larger prediction and goal offsets.
Tactile conditioning generally slows long-horizon error accumulation, but its
effect on task success depends on both the tactile representation and visual
context.
Structured tactile force fields provide the most consistent improvements, with
PointCloud+TacFF achieving the highest overall simulated success rate of
36.1\%.

These trends persist across controlled variations in latent dimensionality,
planning objective, encoder architecture, fusion, and regularization.
We further evaluate them through 900 physical trials across six tasks, two
robot platforms, and three tactile sensing systems.
The physical results reinforce a complementary view of multimodal
representation: explicit geometry is effective when task-relevant structure is
observable, whereas tactile feedback becomes particularly valuable when
contact states are visually ambiguous.

Our contributions are fourfold:
\begin{enumerate}
    \item We introduce \textbf{ContactWorld}, a controlled benchmark for
    studying visual and tactile representations across 12 contact-rich task
    variants within a unified world-model and planning framework.
    We will release the environments, demonstrations, evaluation splits,
    training configurations, and analysis scripts upon publication.

    \item We characterize representation effects through spatial fidelity,
    motion coherence, and predictive stability, relating these properties to
    prediction quality and downstream planning performance.

    \item We quantify how tactile representations affect long-horizon
    prediction and task success across visual modalities, identifying
    structured tactile force fields as the most consistently effective tactile
    representation among those evaluated.

    \item We validate the identified representation trends through controlled
    robustness studies and 900 physical trials spanning six tasks, two robot
    platforms, and three tactile sensing systems.
\end{enumerate}

\section{Related Work}

\begin{figure*}[t]
    \centering
    \includegraphics[width=\textwidth]{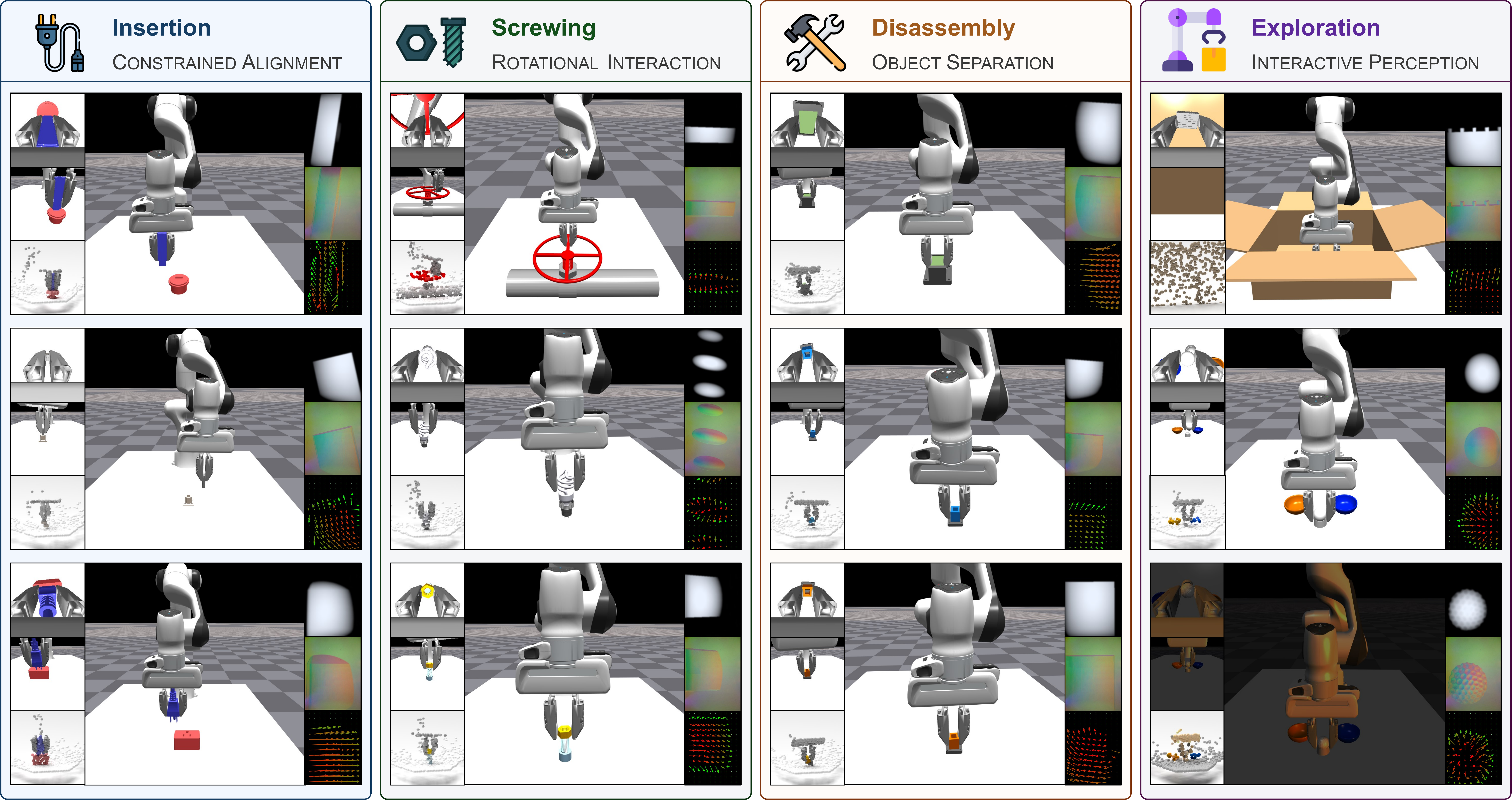}
    \caption{\textbf{ContactWorld benchmark.}
    The benchmark contains 12 task variants across insertion, screwing,
    disassembly, and exploration.
    Each task shows wrist-view RGB, front-view RGB, and PointCloud observations
    alongside TacDepth, TacRGB, and TacFF.
    Disassembly components are colored only for visualization; the simulated
    objects and constraints have similar appearance.}
    \label{fig:teaser}
    \vspace{-10pt}
\end{figure*}

\subsection{World Models for Robotic Manipulation}

World models learn predictive environment dynamics from high-dimensional
observations and support planning, control, and policy learning
~\cite{su2026world}.
Latent world models encode observations into compact states and predict their
future evolution rather than reconstructing raw observations.
Early methods established the effectiveness of latent dynamics for visual
control, while Dreamer and its variants demonstrated policy optimization
through imagined trajectories
~\cite{ha2018worldmodels,hafner2019planet,hafner2020dreamer,
hafner2021dreamerv2,hafner2023dreamerv3}.

Predictive representation-learning methods instead learn dynamics by predicting
abstract embeddings.
I-JEPA and its extensions apply joint-embedding prediction to images, motion,
graphs, and video
~\cite{IJEPA,bardes2023mc,skenderi2023graph,assran2025v}.
Such objectives are attractive for robotics because they can capture
task-relevant dynamics without requiring pixel-level generation
~\cite{ebjepa,he2026demojepa}.
Planning-oriented approaches including DINO-WM and LeWorldModel further learn
action-conditioned dynamics over pretrained representations and optimize
actions directly in latent space
~\cite{dinowm,oquab2023dinov2,maes2026leworldmodel}.
Other work introduces generative, object-centric, or geometry-aware structure
to improve physical prediction
~\cite{bruce2024genie,hu2023gaia1,agarwal2025cosmos,
lin2020gswm,ferraro2025focus,yang2023unisim}.

These studies largely emphasize architectures, training objectives, and
aggregate prediction or control performance
~\cite{wang2026world,terver2025drives}.
ContactWorld instead holds the predictive and planning framework fixed and
examines which observable properties of sensory representations support
spatial reasoning, temporal prediction, and planning during physical contact.

\subsection{Vision--Tactile Learning for Contact-Rich Manipulation}
Tactile sensing provides local measurements of contact, deformation, geometry,
and force that complement visual scene information.
Vision-based sensors such as GelSight and DIGIT, together with array-based and
specialized tactile systems, have enabled learning for insertion, assembly,
grasping, and exploration
~\cite{gelsight,digit,lambeta2024digitizing,huang20243d,
oller2023manipulation,zhang2024gelroller,jiang2025rotipbot}.
Simulation frameworks further support scalable generation of tactile
observations and contact interactions
~\cite{tacsl,si2024difftactile,dang2026hydroshear}.

Recent methods incorporate tactile observations directly into learned robot
policies.
ManiFeel, MimicTouch, ViTacFormer, and related approaches demonstrate their
benefits for imitation learning, multimodal control, and contact-rich
manipulation
~\cite{luu2025manifeel,yu2023mimictouch,heng2025vitacformer,
mao2024multimodal,saka2026contact,liu2025vtdexmanip,
wu2025freetacman,liu2025vitamin}.
These studies also show that tactile utility varies with the task, sensing
configuration, and representation.

Most visuotactile studies assess touch through policy performance, perception,
or sensor modeling, coupling representation choice with the downstream
learning method.
World-model-based planning introduces an additional requirement: visual and
tactile representations must remain informative under action-conditioned
rollouts as prediction errors accumulate.
The Visuo-Tactile World Models framework~\cite{higuera2026visuo}
further shows that jointly predicting visual and tactile futures can improve
contact-rich planning even when candidate actions are evaluated using a
visual objective.
ContactWorld complements this direction through controlled comparisons of
image-, geometry-, deformation-, and force-based representations within a
unified predictive-planning framework.

\section{ContactWorld}
\subsection{Contact-Rich Manipulation Tasks}
ContactWorld is designed to systematically evaluate vision--tactile world models across diverse forms of contact-rich interaction.
As illustrated in Fig.~\ref{fig:teaser}, the benchmark contains four interaction categories comprising 12 task variants: insertion (USB, Peg, and Power Plug), disassembly (Spike Barb, Flat Barb, and Loose Lid), screwing (Bulb, Nut, and Valve), and exploration (Object Search, Normal Sorting, and Dim Sorting).
Together, these tasks span constrained alignment, frictional separation, sustained rotational interaction, and contact-driven perception, providing a controlled testbed for studying multimodal predictive models under distinct contact dynamics.

\textbf{Insertion} requires precise geometric alignment followed by sustained contact during constrained insertion.
The USB, Peg, and Power Plug tasks differ in geometry and contact tolerance, requiring accurate spatial reasoning under partial observability and narrow clearances.
\textbf{Disassembly} requires separating constrained components under frictional contact.
Spike Barb, Flat Barb, and Loose Lid introduce distinct release geometries, directional resistance, and contact transitions, making successful prediction dependent on both interaction state and motion history.
\textbf{Screwing} involves sustained rotational interaction under continuous contact.
Bulb, Nut, and Valve require the model to maintain consistent contact dynamics over longer temporal sequences while reasoning about rotation, alignment, and force transmission.
\textbf{Exploration} emphasizes interactive perception through physical interaction.
Object Search and the two Sorting tasks require the robot to resolve object or environment states that are ambiguous, hidden, or visually degraded, making tactile information particularly relevant for predictive state estimation.

All tasks are implemented in Isaac Gym using TacSL~\cite{tacsl}.
Expert demonstrations are collected at 10~Hz through 6-DoF SpaceMouse teleoperation of a Franka Panda under the randomizations listed in Table~\ref{tab:task_setup}.
Insertion uses a 6-D relative Cartesian pose action with a fixed gripper state,
whereas the other categories additionally include a scalar gripper command.
The resulting dataset contains 1,872 demonstrations and 247,518 interaction
steps.

\subsection{Sensory Modalities and Representation Structure}
\begin{table*}[t]
\centering
\caption{Dataset statistics for the 12 ContactWorld task variants.}
\label{tab:task_setup}
\small
\setlength{\tabcolsep}{5pt}
\renewcommand{\arraystretch}{1.12}

\begin{tabular}{l l c r l}
\toprule
\textbf{Task} & \textbf{Category} & \textbf{Demos} & \textbf{Steps} & \textbf{Randomization} \\
\midrule
USB        & Insertion    & 201 & 13,671 & Plug rot. ($\pm 10^\circ$) + socket pos. ($\pm 0.05$ m) \\
Peg        & Insertion    & 153 & 11,728 & Plug rot. ($\pm 10^\circ$) + socket pos. ($\pm 0.05$ m) \\
Power Plug & Insertion    & 201 & 24,167 & Plug rot. ($\pm 10^\circ$) + socket pos. ($\pm 0.05$ m) \\
\midrule
Spike Barb & Disassembly  & 200 & 13,793 & Socket pos. ($\pm 0.05$ m) \\
Flat Barb  & Disassembly  & 200 & 18,177 & Socket pos. ($\pm 0.05$ m) \\
Loose Lid  & Disassembly  & 200 & 18,042 & Socket pos. ($\pm 0.05$ m) \\
\midrule
Bulb       & Screwing     & 100 & 50,650 & -- \\
Nut        & Screwing     & 101 & 46,164 & -- \\
Valve      & Screwing     & 102 & 16,307 & -- \\
\midrule
Object Search    & Exploration & 108 & 7,844  & Object $x$-position $\{0, 0.05\}$ m \\
Sorting (Normal) & Exploration & 155 & 11,732 & Object placement + lighting \\
Sorting (Dim)    & Exploration & 151 & 15,243 & Object placement + lighting \\
\midrule
\textbf{Total} & -- & \textbf{1,872} & \textbf{247,518} & -- \\
\bottomrule
\end{tabular}
\footnotesize
\parbox{0.72\textwidth}{
\vspace{0.5em}
\textit{Note.}
All tasks use relative Cartesian end-effector control at 10~Hz.
Insertion tasks randomize plug rotation and socket position; disassembly tasks randomize socket position; exploration tasks randomize object position, placement, and/or lighting depending on the task.
``--'' indicates no additional task-level randomization.
}
\end{table*}
ContactWorld provides synchronized visual and tactile observations with different representation structures.
Visual observations include wrist-view and front-view RGB images, as well as point clouds reconstructed from front-view RGB-D observations and subsampled to 1,024 XYZRGB points.
The RGB images retain dense appearance information, whereas the point clouds make depth, 3D scene geometry, and spatial relationships between interacting objects explicit.
The comparison therefore evaluates these observation configurations as provided to the world model, with their differences reflecting both depth availability and representation structure.

For tactile sensing, we consider three representations generated by TacSL: tactile RGB images (TacRGB), tactile depth maps (TacDepth), and tactile force fields (TacFF).
TacRGB captures local tactile appearance, TacDepth represents contact-induced surface deformation, and TacFF encodes spatially distributed three-dimensional force responses, including one normal and two tangential components at each taxel.
Table~\ref{tab:modality_spec} summarizes the observation specifications used throughout the benchmark.
\begin{table}[t]
\centering
\caption{Sensory modalities used in ContactWorld.}
\label{tab:modality_spec}
\footnotesize
\setlength{\tabcolsep}{6pt}
\renewcommand{\arraystretch}{1.12}
\begin{tabular}{llll}
\toprule
Modality & Type & Shape & Encoded information \\
\midrule
Wrist View & RGB & $256\times256\times3$ & Local appearance \\
Front View & RGB & $256\times256\times3$ & Global 2D context \\
PointCloud & XYZRGB & $1024\times6$ & Explicit 3D geometry \\
\midrule
TacRGB & RGB & $320\times240\times3$ & Contact appearance \\
TacDepth & Depth & $320\times240$ & Surface deformation \\
TacFF & Force field & $10\times14\times3$ & Contact force distribution \\
\bottomrule
\end{tabular}
\end{table}

These modalities expose complementary aspects of physical interaction.
Visual observations differ in how explicitly they represent global geometry, whereas tactile observations encode local contact through appearance, deformation, or distributed force.
We quantify these representation differences through three complementary
analyses in Sec.~\ref{sec:analysis}: spatial fidelity and motion coherence for visual representations, and predictive stability for both vision-only and vision--tactile models.
Downstream planning success provides a common task-level evaluation across all modality configurations.

\subsection{World Model Learning and Planning Evaluation}
\label{sec:training_plannig}
\begin{figure*}[t]
    \centering
    \includegraphics[width=1.0\textwidth]{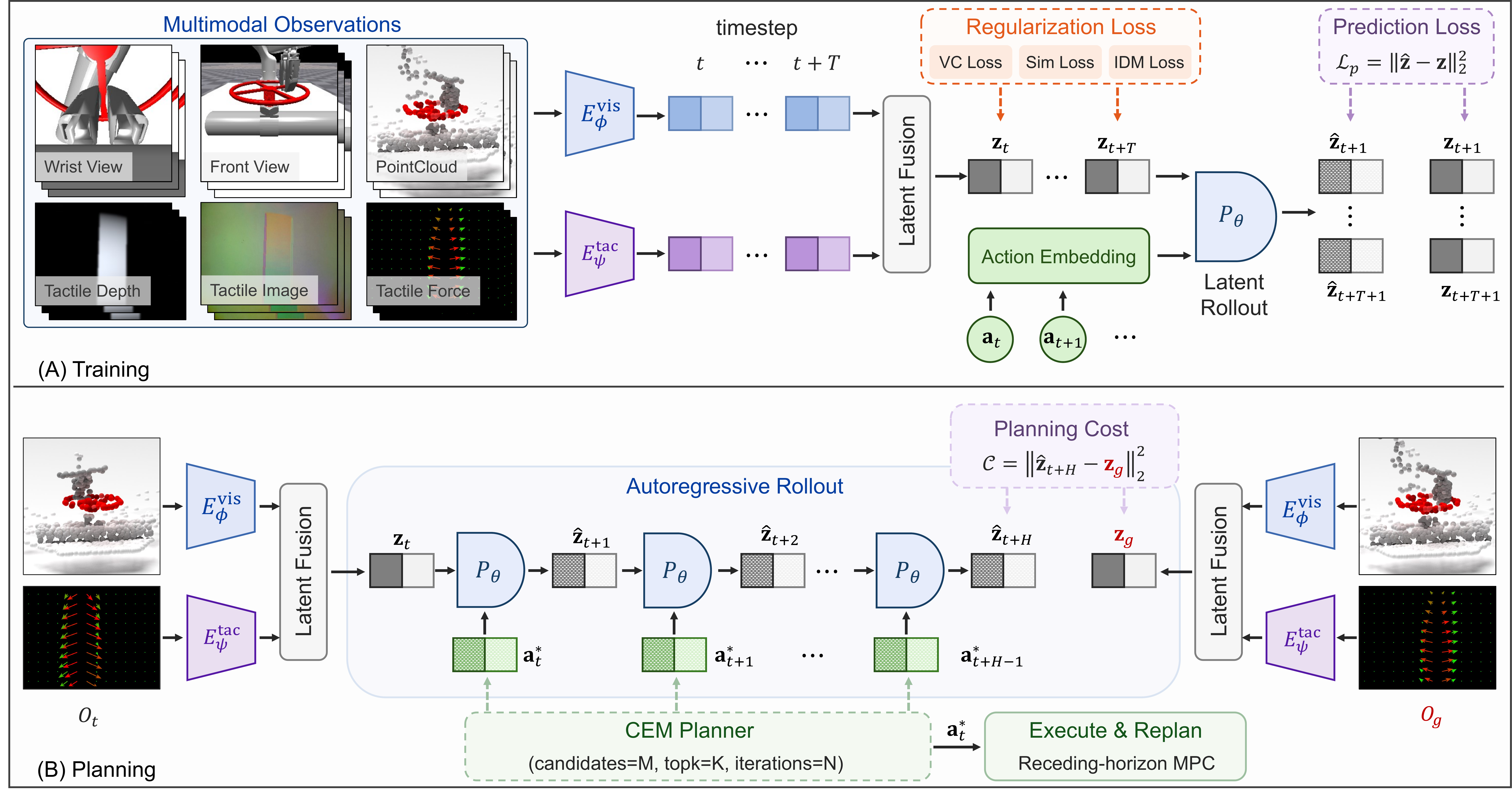}
    \caption{
    \textbf{Training and planning pipeline of ContactWorld.}
    (A) During training, multimodal observations are encoded into latent representations, and the world model learns action-conditioned latent dynamics through autoregressive future prediction and latent regularization.
    (B) During planning, CEM samples candidate action sequences, and the learned world model performs latent rollouts for receding-horizon MPC control.
    }
    \label{fig:architecture}
    \vspace{-10pt}
\end{figure*}
\paragraph{Latent world model}
As illustrated in Fig.~\ref{fig:architecture}(A), ContactWorld adopts a
planning-oriented latent world model based on joint-embedding predictive
learning. Rather than reconstructing future observations, the model learns
action-conditioned dynamics directly in latent space through autoregressive
future prediction. Wrist/front RGB, TacRGB, and TacDepth observations are
encoded using IMPALA-style convolutional encoders~\cite{espeholt2018impala},
while point clouds are processed using a PointNet-style spatial
encoder~\cite{DP3}. TacFF is represented as a low-resolution grid of
three-dimensional force vectors $(f_z,f_x,f_y)$ and encoded using a shared
per-taxel MLP followed by spatial aggregation and latent projection.
All encoders are trained end-to-end from scratch rather than using frozen
pretrained representations.

For multimodal models, visual and tactile embeddings are fused through
latent concatenation and propagated by an action-conditioned GRU
predictor~\cite{ebjepa}.
Given the current latent representation $\mathbf{z}_t$ and an action
sequence $\mathbf{a}_{t:t+K-1}$, the predictor recursively computes
\begin{equation}
    \hat{\mathbf{z}}_{t+k}
    =
    P_\theta(\hat{\mathbf{z}}_{t+k-1},\mathbf{a}_{t+k-1}),
    \qquad k=1,\ldots,K,
\end{equation}
where $\hat{\mathbf{z}}_t=\mathbf{z}_t$.
The prediction objective is
\begin{equation}
    \mathcal{L}_{\mathrm{pred}}
    =
    \frac{1}{K}
    \sum_{k=1}^{K}
    \left\|
    \hat{\mathbf{z}}_{t+k}
    -
    \mathbf{z}_{t+k}
    \right\|_2^2,
\end{equation}

To encourage temporally coherent and action-relevant representations,
we additionally employ temporal similarity and inverse-dynamics
objectives. The temporal similarity loss encourages smooth latent
evolution between adjacent observations:
\begin{equation}
    \mathcal{L}_{\mathrm{sim}}
    =
    \left\|
    \mathbf{z}_{t+1}
    -
    \mathbf{z}_{t}
    \right\|_2^2.
\end{equation}
The inverse-dynamics objective encourages the latent representation to
retain information about the executed action:
\begin{equation}
    \mathcal{L}_{\mathrm{idm}}
    =
    \left\|
    \mathbf{a}_t
    -
    h_\psi(\mathbf{z}_t,\mathbf{z}_{t+1})
    \right\|_2^2,
\end{equation}
where $h_\psi$ denotes an inverse-dynamics predictor that estimates the
executed action from consecutive latent states.

The complete training objective is
\begin{equation}
    \mathcal{L}
    =
    \mathcal{L}_{\mathrm{pred}}
    +
    \lambda_{\mathrm{reg}}\mathcal{L}_{\mathrm{reg}}
    +
    \lambda_{\mathrm{sim}}\mathcal{L}_{\mathrm{sim}}
    +
    \lambda_{\mathrm{idm}}\mathcal{L}_{\mathrm{idm}},
\end{equation}
where $\mathcal{L}_{\mathrm{reg}}$ denotes VICReg-style
variance--covariance regularization~\cite{bardes2022vicreg}.
Unless otherwise specified, this representation regularization is
applied only to the visual latent before multimodal fusion.
The action-conditioned predictor, training objective, and control interface are held fixed across modality configurations.
Modality-specific encoders are required to process the different observation structures, while their outputs use a common per-modality latent dimension.
We separately evaluate dimension-matched multimodal representations and alternative encoder designs to test whether the observed trends arise from latent capacity or encoder choice.

\paragraph{Training configuration}
All models are trained from scratch for 100k gradient steps using AdamW, a batch size of 64, and learning rate and weight decay of $10^{-4}$.
Image and point-cloud encoders output 512-dimensional latents; TacFF uses a 64-dimensional taxel-wise encoding before projection into the shared latent space.
We set the training horizon to $K=2$, $\lambda_{\mathrm{sim}}=\lambda_{\mathrm{idm}}=0.1$, and use unit coefficients for the VICReg variance and covariance terms.

\paragraph{Latent-space planning}
As illustrated in Fig.~\ref{fig:architecture}(B), downstream control is
performed using receding-horizon Model Predictive Control (MPC) with the
Cross-Entropy Method (CEM)~\cite{rubinstein1999cross}.
At each interaction step, the current observation and target observation are
encoded as $\mathbf{z}_t$ and $\mathbf{z}_g$. For each candidate action
sequence, the learned world model performs an autoregressive latent rollout,
and the sequence is evaluated according to the terminal latent-space cost
\begin{equation}
    \mathcal{C}
    =
    \left\|
    \hat{\mathbf{z}}_{t+H}
    -
    \mathbf{z}_g
    \right\|_2^2 .
\end{equation}
CEM samples 100 candidate action sequences over four optimization iterations
and retains the eight lowest-cost elites. The planning rollout horizon is
fixed to six prediction steps. Only the first action of the optimized sequence
is executed before replanning from the newly observed state, yielding
closed-loop control at 10~Hz.

\paragraph{Planning horizons and evaluation}
We distinguish the \emph{model rollout horizon} from the \emph{goal offset}.
The former denotes the number of autoregressive latent prediction steps used
by the planner, whereas the latter denotes the temporal separation between
the initial observation and target observation sampled from a held-out
trajectory. We evaluate goal offsets of 12, 24, 36, and 48 environment steps
using identical initial--goal pairs across modality configurations. Their
corresponding maximum replanning budgets are 15, 30, 45, and 60 interaction
steps, respectively.

Planning success requires joint geometric consistency of both the manipulated
object and robot end-effector with the target state. A rollout is successful
only when the position errors of both are below 1~cm and their orientation
errors are below $15^\circ$. The position threshold corresponds to
approximately 7\% of the overall task-space motion range. Requiring both
object- and end-effector-level consistency prevents partial or unstable
interaction progress from being counted as successful completion. Unless
otherwise specified, each planning configuration is evaluated over 100
parallel environments initialized from randomly sampled held-out trajectory
segments.

\section{What Representation Properties Matter for Contact-Rich World Models?}
\label{sec:analysis}
We systematically evaluate how sensory representation affects prediction and planning across the 12 simulated ContactWorld tasks.
We begin with task-level planning performance across all visual and tactile modality combinations (Table~\ref{tab:main_results}).
We then examine the observed differences through three complementary properties: spatial fidelity and motion coherence for visual representations, and predictive stability for both vision-only and vision--tactile models.
Finally, we analyze the representation-dependent contribution of tactile sensing and conduct controlled studies to test whether the observed trends can be explained by latent dimensionality, planning objective, encoder choice, fusion strategy, or regularization.

\begin{figure*}[t]
    \centering
    \includegraphics[width=0.9\textwidth]{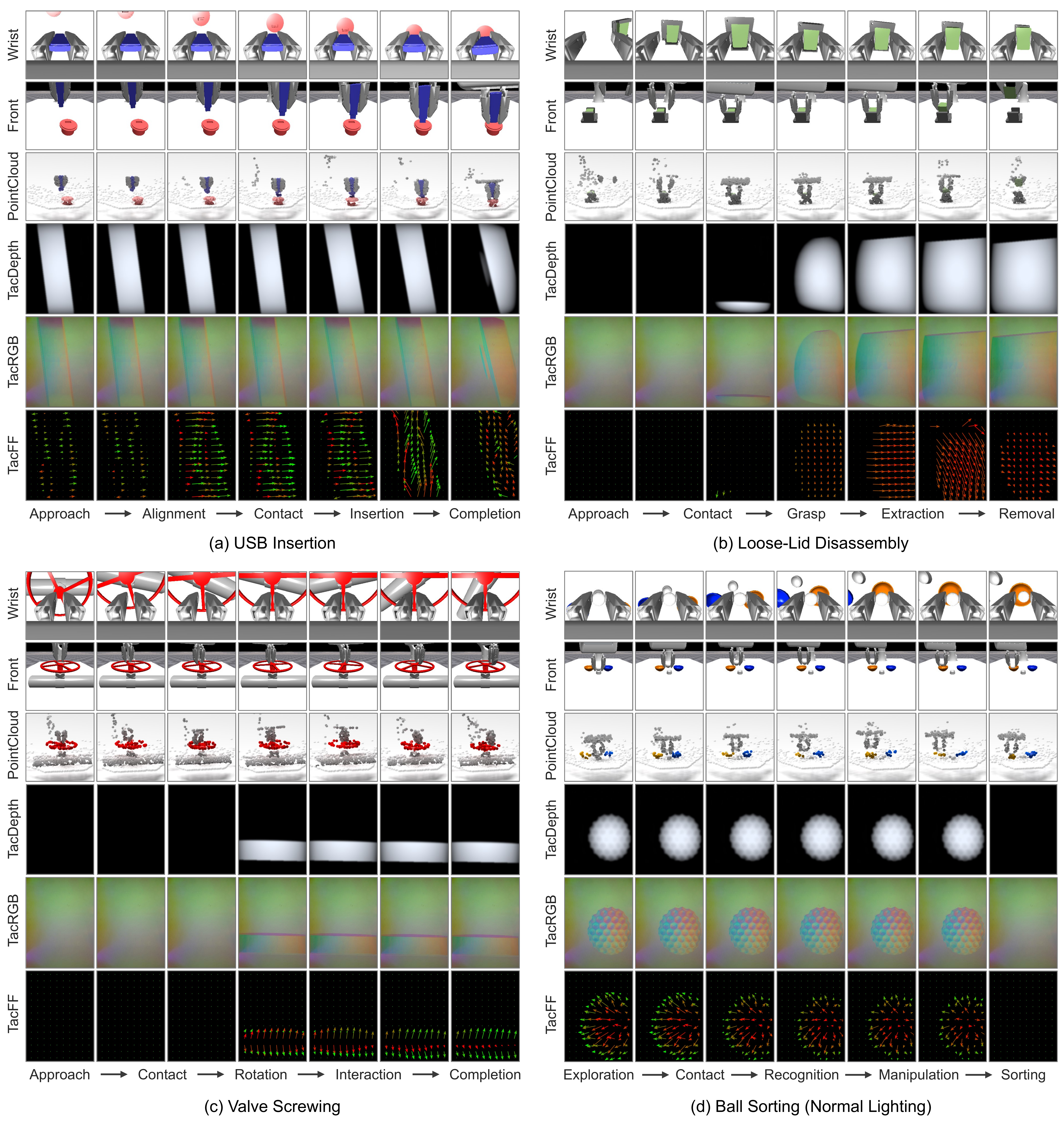}
    \caption{
    \textbf{Temporal multimodal observations across representative ContactWorld tasks.}
    Columns progress from left to right through task-specific interaction phases, while rows show synchronized wrist-view, front-view, point-cloud, TacDepth, TacRGB, and TacFF observations.
    The sequences illustrate viewpoint-dependent occlusion, evolving 3D geometry, and distinct tactile responses across insertion, disassembly, screwing, and exploration.
    }
    \label{fig:time}
\end{figure*}

\begin{table*}[t]
\centering
\caption{
Planning success rates across modality combinations in ContactWorld.
}
\label{tab:main_results}
\scriptsize
\setlength{\tabcolsep}{4.6pt}
\renewcommand{\arraystretch}{1.15}
\begin{tabular}{l|l|rrrr|rrrr|rrrr}
\toprule
Category & Task &
\multicolumn{4}{c|}{Wrist View} &
\multicolumn{4}{c|}{Front View} &
\multicolumn{4}{c}{PointCloud} \\
\cmidrule(lr){3-6}\cmidrule(lr){7-10}\cmidrule(lr){11-14}
& &
Only & +TacDepth & +TacRGB & +TacFF &
Only & +TacDepth & +TacRGB & +TacFF &
Only & +TacDepth & +TacRGB & +TacFF \\
\midrule

\multirow{3}{*}{Insertion}
& USB   & 21.0 & 21.3 & 19.0 & 22.3 & 29.8 & 28.3 & 28.8 & 32.5 & 42.3 & 41.0 & 46.5 & \textbf{50.3} \\
& Peg   & 19.5 & 19.0 & 19.3 & 20.3 & 30.5 & 30.8 & 30.8 & 33.5 & 38.3 & 31.5 & 30.3 & \textbf{40.0} \\
& Power Plug & 38.3 & 36.5 & 38.5 & 37.5 & 46.0 & 46.0 & 40.5 & 46.0 & 60.5 & 55.5 & 56.5 & \textbf{64.0} \\
\midrule

\multirow{3}{*}{Disassembly}
& Spike Barb & 20.8 & 31.5 & 32.8 & 25.5 & 24.0 & 18.8 & 22.3 & 23.3 & 53.3 & 42.3 & 39.8 & \textbf{56.5} \\
& Flat Barb & 19.3 & 23.8 & 25.3 & 22.3 & 13.5 & 15.5 & 13.5 & 10.5 & 35.0 & 34.3 & 34.3 & \textbf{41.3} \\
& Loose Lid   & 23.3 & 19.5 & 25.3 & 23.3 & 13.0 & 13.8 & 11.0 & 14.3 & 15.8 & 18.0 & 30.5 & \textbf{37.5} \\
\midrule

\multirow{3}{*}{Screwing}
& Nut   & 17.0 & 16.3 & 18.3 & 19.3 & 18.3 & 19.0 & 21.0 & 20.0 & 20.3 & 24.8 & 21.8 & \textbf{25.0} \\
& Bulb  & 10.3 & 11.3 & 13.3 & 13.0 & 10.3 & 13.5 & 10.3 & 10.0 & \textbf{15.5} & 14.5 & 14.3 & \textbf{15.5} \\
& Valve & 40.8 & \textbf{59.8} & 36.8 & 51.3 & 40.8 & 56.5 & 55.5 & 42.8 & 51.5 & 41.8 & 44.8 & 52.5 \\
\midrule

\multirow{3}{*}{Exploration}
& Object Search & 10.5 & 9.5  & \textbf{12.5} & 12.0 & 9.0  & 9.5  & 8.8  & 8.0  & 9.0  & 9.8  & 9.8  & 8.0 \\
& Normal    & 16.0 & 27.5 & 24.0 & 19.0 & 15.5 & 29.0 & 22.5 & 15.3 & 27.0 & \textbf{33.0} & 24.5 & 24.0 \\
& Dim       & 12.3 & 15.5 & 14.0 & 13.8 & 13.5 & 14.0 & 11.5 & 13.3 & 16.8 & 15.8 & 15.3 & \textbf{18.3} \\
\midrule

\multicolumn{2}{c|}{Average}
& 20.7 & 24.3 & 23.2 & 23.3
& 22.0 & 24.5 & 23.0 & 22.4
& 32.1 & 30.2 & 30.7 & \textbf{36.1} \\

\bottomrule
\end{tabular}
\parbox{0.94\linewidth}{
\vspace{4pt}
\footnotesize
Each entry reports the average success rate across goal offsets
of 12, 24, 36, and 48 steps, with 100 evaluation rollouts per
offset for each task--modality combination.
Bold values indicate the best-performing modality combination for each task.
}
\end{table*}

\subsection{Overall Planning Performance Across Representations}
\label{sec:overall_performance}
Table~\ref{tab:main_results} reports planning success across all visual and tactile modality combinations.
The results reveal substantial differences across visual representations.
Averaged over the 12 tasks, wrist-view and front-view RGB achieve 20.7\% and 22.0\% vision-only success, respectively, whereas point-cloud observations increase the average success rate to 32.1\%.
The point-cloud advantage is particularly pronounced in insertion and disassembly, where planning requires accurate reasoning about constrained geometry and evolving object--environment relationships.

The contribution of tactile sensing is more representation- and task-dependent.
TacDepth provides the largest average improvement for wrist- and front-view models, while TacFF provides the largest improvement when combined with point clouds.
PointCloud+TacFF achieves the highest overall simulated success rate of 36.1\%.
However, no tactile representation improves every visual modality and task, indicating that tactile information should be evaluated through both its predictive contribution and its effect on downstream planning.

Fig.~\ref{fig:time} provides qualitative context for these task-level results through synchronized observation sequences from the four interaction categories.
Wrist-view observations provide close-range interaction details, although partial occlusion occurs in some phases and tasks.
Front-view observations provide broader scene context, while point clouds explicitly encode the evolving 3D configurations of the robot, manipulated object, and surrounding environment.

The tactile sequences illustrate how the same physical interaction is encoded in different forms.
TacDepth captures contact-induced surface deformation, TacRGB records changes in tactile appearance, and TacFF represents the spatial distribution of normal and tangential forces.
Their responses vary across interaction phases and task categories, providing qualitative evidence that the tactile representations expose different aspects of contact.
We next quantify the visual differences through spatial fidelity and motion coherence, followed by predictive-stability and task-level analyses of the vision--tactile models.

\subsection{Visual Representation Properties}
\label{sec:visual_properties}
Task success identifies which visual representations perform better, but does not explain which representation properties support predictive planning.
We therefore evaluate two interpretable properties of the learned visual latents using simulator state information: \emph{spatial fidelity} and \emph{motion coherence}.
These diagnostics are evaluated on episode-disjoint trajectories and are computed independently of planning success.

\paragraph{Spatial fidelity}
Spatial fidelity measures whether a learned representation retains task-relevant geometric relations.
For each visual modality, we fit a linear ridge probe from the frozen latent representation $\mathbf{z}_t$ to a physical relation
$\mathbf{y}_t$:
\begin{equation}
    \hat{\mathbf{y}}_t
    =
    \mathbf{W}\mathbf{z}_t+\mathbf{b},
\end{equation}
where $\mathbf{y}_t$ denotes either the end-effector--target or object--target relative position.
The probe is trained on training episodes, with its regularization coefficient selected using validation episodes, and performance is reported on held-out evaluation episodes.

We report the normalized root mean squared error (NRMSE), defined as the
probe RMSE normalized by the error of a constant predictor that outputs the
training-set mean:
\begin{equation}
    \operatorname{NRMSE}
    =
    \frac{
    \sqrt{\frac{1}{N}\sum_{t=1}^{N}
    \left\|\hat{\mathbf{y}}_t-\mathbf{y}_t\right\|_2^2}
    }{
    \sqrt{\frac{1}{N}\sum_{t=1}^{N}
    \left\|\bar{\mathbf{y}}_{\mathrm{train}}-\mathbf{y}_t\right\|_2^2}
    }.
    \label{eq:spatial_fidelity}
\end{equation}
Lower NRMSE indicates that the latent representation preserves the corresponding spatial relation more accurately; a value below one indicates improvement over the constant baseline.

\begin{table}[t]
\centering
\caption{\textbf{Spatial fidelity of visual representations.}}
\label{tab:spatial_fidelity}

\renewcommand{\arraystretch}{1.12}
\setlength{\tabcolsep}{4pt}

\begin{tabularx}{\columnwidth}{
    @{}
    >{\raggedright\arraybackslash}X
    >{\centering\arraybackslash}p{0.29\columnwidth}
    >{\centering\arraybackslash}p{0.34\columnwidth}
    @{}
}
\toprule
\textbf{Representation} &
\shortstack{\textbf{EEF--Target}\\\textbf{NRMSE} $\downarrow$} &
\shortstack{\textbf{Object--Target}\\\textbf{NRMSE} $\downarrow$} \\
\midrule
Wrist View & 0.316 & 0.347 \\
Front View & 1.500 & 1.620 \\
PointCloud & \textbf{0.263} & \textbf{0.213} \\
\bottomrule
\end{tabularx}

\vspace{3pt}
\parbox{\columnwidth}{
\footnotesize
Values report median normalized probe error across applicable tasks; lower is better.
EEF--target results are aggregated over all 12 tasks.
Object--target results are aggregated over the nine tasks for which the
corresponding task-relevant relation is defined.
}
\end{table}

As shown in Table~\ref{tab:spatial_fidelity}, point-cloud latents achieve the lowest normalized errors for both spatial relations.
Although point clouds explicitly encode 3D coordinates at the input, this result is not guaranteed after representation learning: the probe evaluates whether task-relevant relative geometry remains recoverable from the learned latent used for prediction and planning.
Wrist-view latents retain these local spatial relations more accurately than front-view latents, while point-cloud latents provide the strongest overall spatial fidelity.
This result is consistent with the strong point-cloud performance in geometrically constrained insertion and disassembly tasks.

\begin{table}[t]
\centering
\caption{
\textbf{Motion coherence of visual representations.}
}
\label{tab:motion_coherence}
\small
\setlength{\tabcolsep}{9pt}
\renewcommand{\arraystretch}{1.12}
\begin{tabular}{lcc}
\toprule
\textbf{Representation} &
\shortstack{\textbf{Motion Alignment}\\
$\rho_{\mathrm{motion}}\uparrow$} &
\shortstack{\textbf{Motion Contrast}\\
$C_{\mathrm{motion}}\uparrow$} \\
\midrule
Wrist View  & 0.304 & \textbf{1.637} \\
Front View  & 0.321 & 1.541 \\
PointCloud  & \textbf{0.405} & 1.488 \\
\bottomrule
\end{tabular}

\parbox{1.0\linewidth}{
\vspace{5pt}
\footnotesize
Each task value is first aggregated across evaluation episodes, and the table
reports the median across the 12 tasks.
Higher $\rho_{\mathrm{motion}}$ indicates more consistent correspondence
between physical and latent motion, while higher $C_{\mathrm{motion}}$
indicates stronger discrimination between large and small physical motions.
The front-view contrast is aggregated over the 11 tasks for which the statistic
is defined.
}
\end{table}

\begin{figure*}[t]
    \centering
    \includegraphics[width=0.8\textwidth]{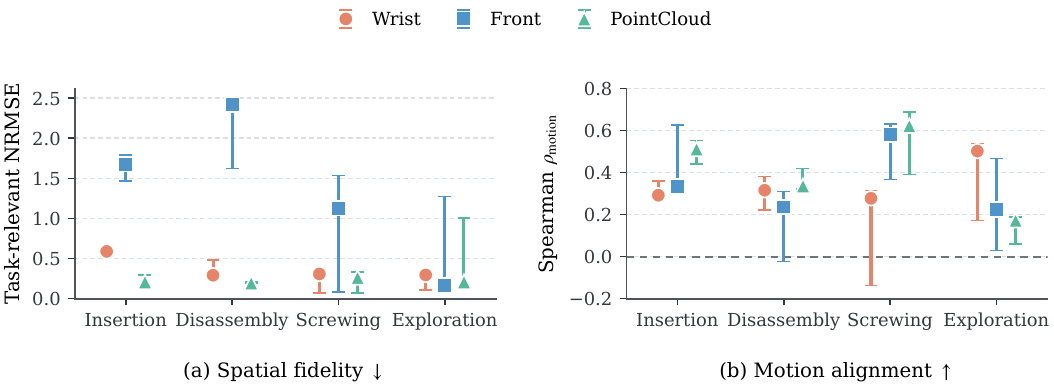}
\caption{
\textbf{Visual representation properties across interaction categories.}
Markers indicate the median task-level result within each category, and whiskers span the minimum and maximum across its three tasks.
(a) Spatial fidelity measures the preservation of task-relevant physical relations using normalized reconstruction error; lower is better.
(b) Motion alignment measures the Spearman correlation between physical displacement and latent change; higher is better, while the dashed line denotes zero correlation.
}
    \label{fig:representation_properties}
\end{figure*}

\paragraph{Motion coherence}
Spatial fidelity evaluates geometric information at individual timesteps, whereas predictive planning also requires representational changes to follow the physical evolution of the interaction.
For consecutive observations, we define the dimension-normalized latent change as
\begin{equation}
    \Delta z_t
    =
    \frac{
    \left\|\mathbf{z}_{t+1}-\mathbf{z}_{t}\right\|_2
    }{\sqrt{D}},
    \label{eq:latent_motion}
\end{equation}
where $D$ is the latent dimensionality.
The corresponding physical displacement $\Delta s_t$ is constructed from task-relevant spatial relations, including target-relative translation, rotational motion, and object--end-effector displacement.
Each component is normalized using a robust motion scale estimated from the training episodes.

We report two complementary metrics.
\emph{Motion alignment} measures the monotonic correspondence between physical and latent motion using Spearman rank correlation:
\begin{equation}
    \rho_{\mathrm{motion}}
    =
    \operatorname{Spearman}
    \left(\Delta s_t,\Delta z_t\right).
    \label{eq:motion_alignment}
\end{equation}
A higher value indicates that larger physical displacements more consistently produce larger representational changes, without assuming a linear relationship.

\emph{Motion contrast} measures whether the representation distinguishes large physical motions from small ones:
\begin{equation}
    C_{\mathrm{motion}}
    =
    \frac{
    \operatorname{median}
    \left(\Delta z_t\mid\Delta s_t\geq Q_{75}\right)
    }{
    \operatorname{median}
    \left(\Delta z_t\mid\Delta s_t\leq Q_{25}\right)
    },
    \label{eq:motion_contrast}
\end{equation}
where $Q_{25}$ and $Q_{75}$ denote the lower and upper quartiles of physical displacement within an episode.
Both metrics are computed separately for each evaluation episode, aggregated to the task level by the median, and then summarized across tasks.

Table~\ref{tab:motion_coherence} shows that point-cloud representations achieve the highest overall motion alignment.
Their latent changes therefore maintain the most consistent monotonic correspondence with task-relevant physical motion across the evaluated tasks.
Wrist-view representations instead achieve the highest motion contrast, reflecting stronger sensitivity to large local interaction changes that occupy a substantial portion of the wrist image.
Front-view representations fall between the two in overall alignment, while showing strong contrast during sustained rotational interactions.

These results distinguish motion correspondence from motion sensitivity.
Wrist observations can respond strongly to salient local changes, but their response is also affected by the motion of the camera and changes in local visibility.
Point clouds do not produce the largest contrast in every interaction category, but provide more consistent alignment between representational and physical motion.
Together with their generally stronger spatial fidelity, this consistency helps explain their higher average planning performance, while the category-level analysis below reveals important interaction-dependent exceptions.

Figure~\ref{fig:representation_properties} further decomposes spatial fidelity and motion alignment across the four interaction categories.
The category-level results reveal that the aggregate advantage of point clouds is interaction-dependent rather than universal.
For insertion and disassembly, point clouds combine low spatial error with the strongest or comparable motion alignment, consistent with the importance of preserving constrained 3D relations among the end effector, object, and target.
For screwing, point clouds and front-view RGB exhibit similarly strong alignment, suggesting that sustained rotational motion remains observable from both explicit geometry and a stable external viewpoint.

Exploration exhibits a different pattern.
Although front-view RGB and point clouds achieve relatively low median spatial error, their task-level variation is substantial, reflecting differences in visibility across search and sorting tasks.
Wrist-view RGB instead provides the strongest motion alignment, particularly because local object and gripper motion can remain visible from the wrist camera when the interaction is less observable from the external view.
Thus, no representation dominates every interaction category: point clouds are most reliable when planning depends on global geometry and constrained scene evolution, whereas wrist observations can better expose local motion under external occlusion.

\begin{table*}[t]
    \centering
    \caption{
    \textbf{Predictive stability under autoregressive latent rollout.}
    }
    \label{tab:predictive_stability}
    \setlength{\tabcolsep}{5.5pt}
    \renewcommand{\arraystretch}{1.12}
    \begin{tabular}{lcccccccc}
        \toprule
        & \multicolumn{4}{c}{Growth Ratio $G_{6/1}$ $\downarrow$}
        & \multicolumn{4}{c}{Growth Rate $k$ $\downarrow$} \\
        \cmidrule(lr){2-5}
        \cmidrule(lr){6-9}
        Visual Representation
        & Only & +TacDepth & +TacRGB & +TacFF
        & Only & +TacDepth & +TacRGB & +TacFF \\
        \midrule
        Wrist
        & 12.86 & 10.48 & \textbf{9.66} & 12.45
        & 0.503 & 0.469 & \textbf{0.458} & 0.499 \\

        Front
        & 11.74 & \textbf{9.06} & 9.29 & 10.56
        & 0.477 & 0.438 & \textbf{0.436} & 0.469 \\

        PointCloud
        & 4.55 & 4.34 & \textbf{3.85} & 4.46
        & 0.305 & 0.316 & \textbf{0.295} & 0.303 \\
        \bottomrule
    \end{tabular}
\parbox{0.76\linewidth}{
\vspace{5pt}
\footnotesize
We report the median growth ratio $G_{6/1}=e_6/e_1$ and fitted exponential error-growth rate $k$ across 12 tasks.
Lower values indicate slower relative error accumulation.
The best tactile configuration for each visual representation is shown in bold.
}
\end{table*}
\subsection{Predictive Stability Under Extended Horizons}
\label{sec:predictive_stability}
We next examine how representation choice affects prediction and planning as the temporal horizon increases.
Unlike spatial fidelity and motion coherence, which characterize individual observations or consecutive observation changes, predictive stability measures how rapidly errors accumulate during autoregressive latent rollout.
It is evaluated for both vision-only and vision--tactile models.

Let $e_h$ represent the latent prediction MSE at rollout horizon $h$.
We summarize error accumulation using the growth ratio
\begin{equation}
    G_{H/1} = \frac{e_H}{e_1},
    \label{eq:growth_ratio}
\end{equation}
where $H=6$ is the longest evaluated rollout horizon.
We additionally fit an exponential growth model,
\begin{equation}
    e_h \approx e_1 \exp\!\left(k(h-1)\right),
    \label{eq:error_growth}
\end{equation}
where $k$ denotes the fitted error-growth rate.
Lower values of both $G_{H/1}$ and $k$ indicate slower relative error accumulation.
Because these measures characterize error growth rather than absolute latent MSE, they reduce the influence of representation-dependent latent scales.

\begin{figure*}[t]
    \centering
    \includegraphics[width=1.0\textwidth]{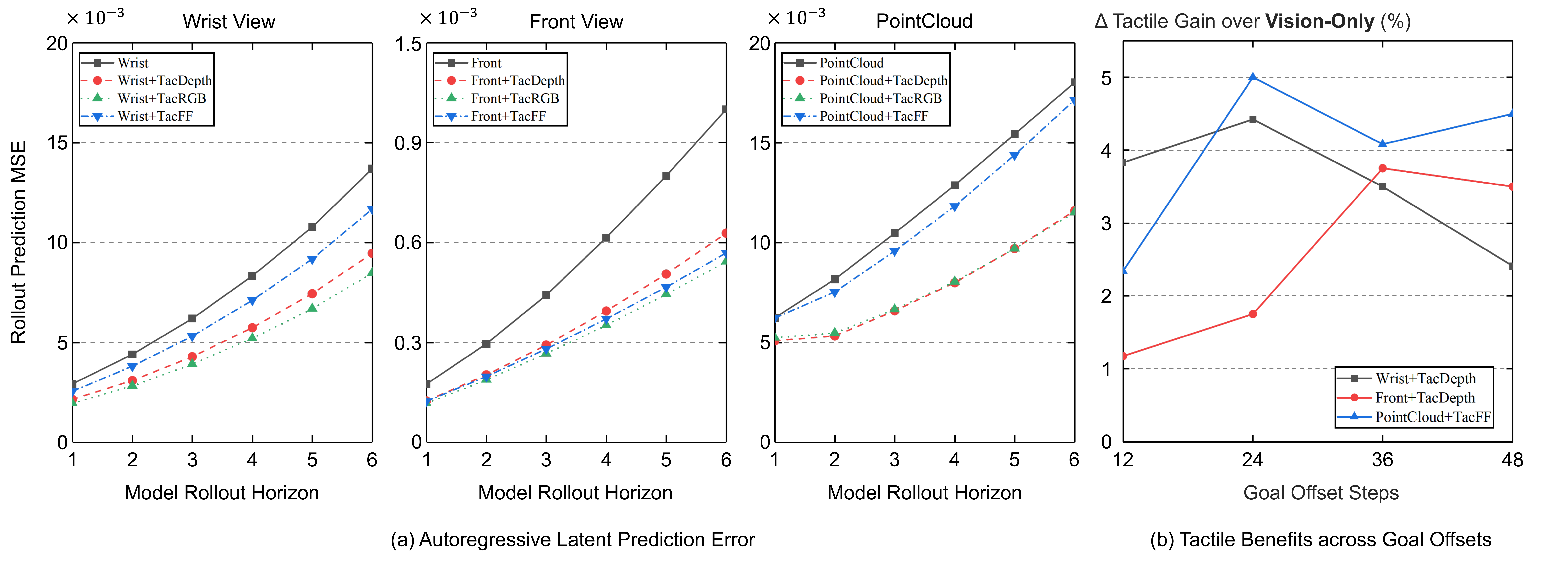}
\caption{
\textbf{Predictive robustness under increasing temporal horizons.}
(a) Autoregressive latent prediction error on the representative USB
insertion task as the model rollout horizon increases.
Tactile-conditioned models generally exhibit slower error accumulation than their vision-only counterparts.
(b) Tactile gains over vision-only baselines across increasing goal offsets.
Together, these results connect improved multistep latent prediction with more robust planning under increased temporal demands.
}
    \label{fig:robustness}
\end{figure*}

Table~\ref{tab:predictive_stability} reports the median results across all 12 tasks.
Point-cloud models exhibit substantially slower relative error growth than the image-based models.
Tactile observations reduce the growth ratio across all three visual representations, although their effects on the fitted growth rate are not uniform.
TacRGB provides the strongest predictive stability for wrist-view and point-cloud models, while TacDepth and TacRGB perform similarly with the front view.
Figure~\ref{fig:robustness}(a) provides a representative visualization of autoregressive error growth on the USB insertion task

These results show that representation choice substantially affects error accumulation during autoregressive prediction, and that tactile observations can improve multistep predictive stability.
Predictive stability explains one component of long-horizon planning performance, but does not alone determine the final task outcome.
For example, TacRGB yields the lowest relative error growth with point clouds, whereas PointCloud+TacFF achieves the highest planning success.
This comparison indicates that reliable planning depends on both stable prediction and the task relevance of the information preserved by the representation.

We further evaluate planning under increasing temporal demand by varying the goal offset from 12 to 48 steps.
As shown in Table~\ref{tab:long_horizon}, success decreases for every configuration as the target state moves farther into the future, while point clouds remain the strongest visual representation at every evaluated offset.
PointCloud+TacFF consistently improves over PointCloud-only, with gains of 2.3, 5.0, 4.1, and 4.5 percentage points at offsets of 12, 24, 36, and 48 steps, respectively.
Figure~\ref{fig:robustness}(b) summarizes the corresponding tactile gains for the selected visual--tactile pairings.

Together, these results show that representation choice affects both autoregressive prediction and planning under extended temporal demands.
Point clouds provide consistently slower relative error accumulation and stronger long-offset planning, while tactile observations can further improve prediction or planning depending on how contact information is represented.

\subsection{Representation-Dependent Benefits of Tactile Sensing}
\label{sec:tactile_benefits}
We next examine how the downstream benefit of tactile sensing depends on its representation and the accompanying visual observation.
As shown in Fig.~\ref{fig:performance}(a), TacDepth provides the largest average improvement over the corresponding vision-only baseline for wrist- and front-view models, whereas point-cloud models benefit most from TacFF.
Among all evaluated configurations, PointCloud+TacFF achieves the highest average simulated success rate of 36.1\%.

These results show that the contribution of touch cannot be characterized independently of the visual representation with which it is combined.
For image-based models, tactile deformation provides local contact information that may compensate for visual ambiguity near the interaction region.
Point clouds already encode explicit scene geometry, whereas TacFF introduces structured contact-motion information that is not directly available from geometric observations.
This complementarity provides a possible explanation for the particularly strong performance of PointCloud+TacFF.

The category-level results in Fig.~\ref{fig:performance}(b) further show that no tactile representation is uniformly optimal across interaction types.
TacFF provides consistent improvements in insertion and substantial gains in disassembly, while TacDepth produces larger gains in several screwing and exploration settings.
These patterns are consistent with the different information exposed by the two representations: TacFF explicitly represents directional contact response, whereas TacDepth captures local surface deformation.
Therefore, the utility of tactile sensing depends not only on adding touch, but also on whether its representation provides information that is complementary to the visual observation and relevant to the interaction.

\begin{figure*}[t]
    \centering
    \includegraphics[width=1.0\textwidth]{figures/performance.pdf}
    \caption{
    \textbf{Representation effects on predictive planning.}
    (a) Planning performance landscape across visual--tactile representations,
    showing average success rate, planning frequency, and model size (bubble area).
    (b) Interaction-dependent tactile benefits over the corresponding vision-only
    baselines. Points denote individual task results; boxes indicate median and
    interquartile ranges.
    }
    \label{fig:performance}    
\end{figure*}

\begin{table*}[t]
\centering
\caption{
Planning performance under increasing goal offsets.
}
\label{tab:long_horizon}
\scriptsize
\setlength{\tabcolsep}{6.0pt}
\renewcommand{\arraystretch}{1.18}
\begin{tabular}{c|rrrr|rrrr|rrrr}
\toprule
\multirow{2}{*}{Offset Steps} &
\multicolumn{4}{c|}{Wrist View} &
\multicolumn{4}{c|}{Front View} &
\multicolumn{4}{c}{PointCloud} \\
\cmidrule(lr){2-5}\cmidrule(lr){6-9}\cmidrule(lr){10-13}
&
Only & +TacDepth & +TacRGB & +TacFF &
Only & +TacDepth & +TacRGB & +TacFF &
Only & +TacDepth & +TacRGB & +TacFF \\
\midrule
12 & 41.4 & 45.3 & 45.5 & 44.7
   & 44.8 & 45.9 & 44.2 & 45.3
   & 52.1 & 51.7 & 51.0 & \textbf{54.4} \\

24 & 21.1 & 25.5 & 25.6 & 26.2
   & 23.7 & 25.4 & 25.3 & 23.8
   & 36.6 & 34.3 & 36.2 & \textbf{41.6} \\

36 & 12.3 & 15.8 & 13.3 & 14.3
   & 11.7 & 15.4 & 13.7 & 12.9
   & 23.7 & 21.3 & 20.8 & \textbf{27.8} \\

48 & 8.2 & 10.6 & 8.6 & 8.0
   & 7.9 & 11.4 & 9.0 & 7.8
   & 16.0 & 13.5 & 14.8 & \textbf{20.5} \\
\bottomrule
\end{tabular}

\parbox{0.92\linewidth}{
\vspace{4pt}
\footnotesize
Each entry reports the average success rate over 100 evaluation trials across all 12 tasks at a fixed goal-offset step.
Larger offsets correspond to longer-term prediction and planning objectives.}
\end{table*}

\subsection{Controlled Ablations and Robustness Analysis}
\label{sec:controlled_representation}
The preceding results establish consistent associations between sensory representation, predictive behavior, and planning success.
We next examine whether the main trends remain under controlled changes to the planning objective, latent dimensionality, planning horizon, encoder architecture, fusion mechanism, and representation regularization.
These experiments are intended to evaluate the robustness of the observed trends.

\paragraph{Planning objective and latent dimensionality}
Adding a tactile branch changes both the information available to the world model and the dimensionality and composition of the latent planning objective.
We therefore evaluate two controlled variants on one representative task from each interaction category, as reported in Table~\ref{tab:controlled_ablation}.

\emph{Vis.-Cost} retains PointCloud+TacFF as the predictive-model input but evaluates candidate action sequences using only the predicted visual latent, matching the form of the PointCloud-only planning cost.
This setting is similar in spirit to Visuo-Tactile World Models~\cite{higuera2026visuo}, in which tactile observations contribute to predictive dynamics while planning uses a visual objective.
\emph{Dim.-Match} retains both modalities while matching the total latent dimensionality of the PointCloud-only model.

PointCloud+TacFF improves the four-task average success rate from 32.3\% to 39.5\%.
Most of this improvement is retained with the visual-only planning cost (38.6\%), indicating that the gain does not arise solely from adding a tactile term to the planning objective.
The dimension-matched variant also improves over PointCloud-only, achieving 36.5\%.
Although the magnitude varies across tasks, these results indicate that the overall multimodal gain cannot be explained solely by the planning-cost composition or increased latent capacity.

\begin{figure}[t]
  \centering
  \includegraphics[width=1.0\linewidth]{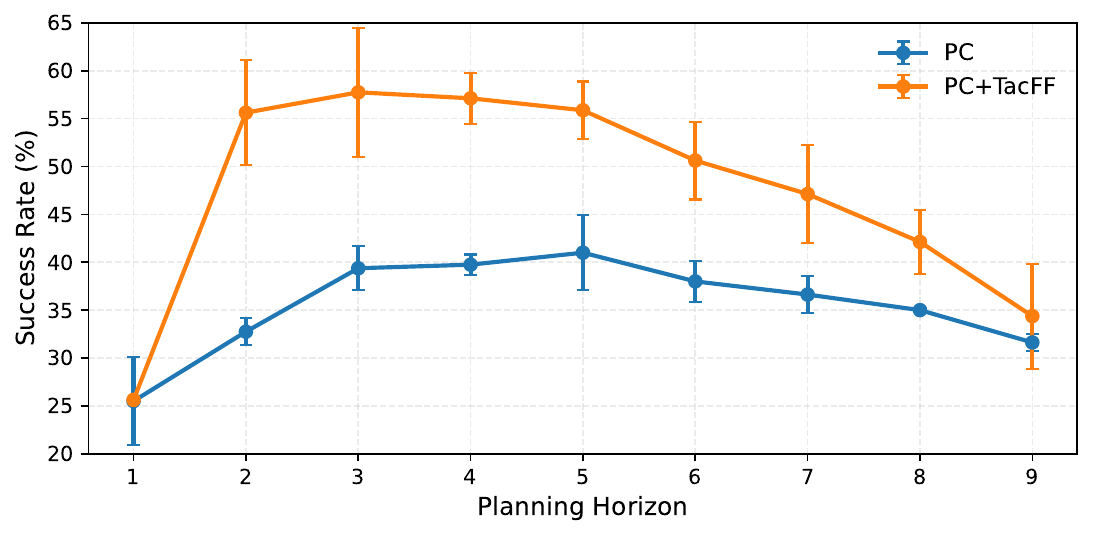}
  \parbox{0.9\linewidth}{
    \caption{
    \textbf{Robustness to CEM planning horizon.}
    Success rates of PointCloud (PC) and PointCloud+TacFF as the CEM horizon varies from 1 to 9 steps.
    Results show mean$\pm$std over three seeds with a fixed world-model training horizon of $H_{\mathrm{train}}=2$.
    }
  \label{fig:planning_horizon}}
\end{figure}

\paragraph{Planning-horizon robustness}
We next vary the CEM planning horizon from 1 to 9 while keeping the trained world model and all other planning settings fixed (Fig.~\ref{fig:planning_horizon}).
At a one-step horizon, PointCloud and PointCloud+TacFF achieve similar performance.
The advantage of PointCloud+TacFF becomes more apparent at intermediate horizons, where action selection depends on predicted interaction evolution over multiple steps.
At the longest horizons, performance decreases for both models, consistent with the increasing difficulty of optimizing actions through longer autoregressive predictions.
The relative advantage of PointCloud+TacFF across multiple planning horizons
shows that its improvement is not restricted to a single CEM horizon setting.
However, the eventual degradation of both models also highlights the remaining
limitations of long-horizon latent rollout.

\begin{table}[t]
\centering
\caption{
\textbf{Controlled validation of representation effects across four interaction categories.}
}
\label{tab:controlled_ablation}
\setlength{\tabcolsep}{4pt}
\renewcommand{\arraystretch}{1.05}

\resizebox{\columnwidth}{!}{%
\begin{tabular}{llcccc}
\toprule
\textbf{Category} &
\textbf{Task} &
\textbf{PC} &
\textbf{PC+TacFF} &
\shortstack{\textbf{PC+TacFF}\\\textbf{(Vis.-Cost)}} &
\shortstack{\textbf{PC+TacFF}\\\textbf{(Dim.-Match)}} \\
\midrule
Insertion
& USB
& $39.3{\pm}0.4$
& $46.8{\pm}1.1$
& $47.8{\pm}1.5$
& $40.3{\pm}3.5$ \\
Disassembly
& Loose Lid
& $21.3{\pm}2.2$
& $35.3{\pm}2.6$
& $36.9{\pm}0.6$
& $31.2{\pm}2.8$ \\
Screwing
& Valve
& $53.6{\pm}1.6$
& $60.8{\pm}1.5$
& $53.3{\pm}1.8$
& $65.3{\pm}0.9$ \\
Exploration
& Dim
& $14.9{\pm}1.5$
& $15.1{\pm}2.0$
& $16.3{\pm}1.9$
& $9.5{\pm}0.7$ \\
\midrule
\textbf{Average}
& {}
& 32.3
& \textbf{39.5}
& 38.6
& 36.5 \\
\bottomrule
\end{tabular}%
}

\vspace{3pt}
\parbox{1.0\linewidth}{
{\scriptsize
Mean$\pm$std over three seeds and four goal offsets (12/24/36/48), totaling 1,200 rollouts per task--method pair with episode-disjoint splits.
Vis.-Cost uses a visual-only planning objective; Dim.-Match matches latent dimensionality.
}}
\end{table}
\begin{figure*}[t]
    \centering
    \includegraphics[width=1.0\textwidth]{figures/vis_tac_encoder.pdf}
    \caption{
    \textbf{Visual and tactile encoder ablations on USB insertion.}
    (a) Task-trained IMPALA, frozen DINOv2, and train-from-scratch ViT encoders
    for image-based visual observations.
    (b) Task-trained IMPALA, AnyTouch~2, and Sparsh encoders for TacRGB.
    All encoders are compared within the same world-model and planning framework.
    }
    \label{fig:vis_tac_encoder}
\end{figure*}

\paragraph{Encoder choice}
We further test whether the observed modality trends depend on the sensory encoder used within the world model (Fig.~\ref{fig:vis_tac_encoder}).
For image-based visual observations, the task-trained IMPALA encoder consistently outperforms frozen DINOv2~\cite{oquab2023dinov2} and a train-from-scratch ViT~\cite{terver2025drives} across the evaluated tactile configurations.
This result is consistent with the observation in LeWorldModel~\cite{maes2026leworldmodel} that representations developed for semantic visual perception are not necessarily optimal for action-conditioned latent prediction and planning.

We observe a similar dependence on tactile encoder choice.
Pretrained AnyTouch~2~\cite{anytouch2} and Sparsh~\cite{sparsh} are competitive in several image-based configurations, whereas the task-trained IMPALA encoder performs substantially better when TacRGB is paired with PointCloud.
These comparisons show that the main results are not explained simply by the use of a particular pretrained encoder.
They also suggest that encoder suitability for world-model planning depends on whether the resulting latent features preserve predictable interaction information, rather than on perceptual pretraining performance alone.

\begin{figure*}[t]
    \centering
    \includegraphics[width=0.9\textwidth]{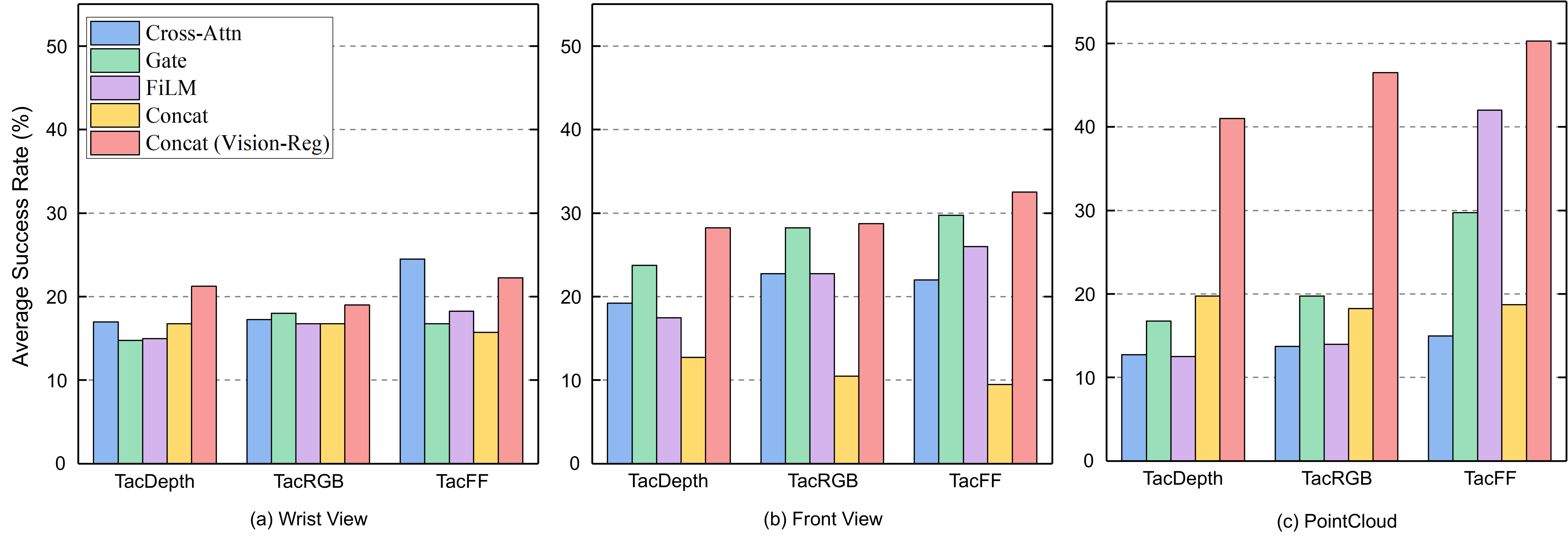}
    \caption{
    \textbf{Fusion and modality-specific regularization ablations on USB
    insertion.}
    We compare cross-attention, gated fusion, FiLM, late concatenation, and late
    concatenation with visual-only regularization for
    (a) wrist-view, (b) front-view, and (c) point-cloud observations.
    Bars report average planning success across goal offsets of 12, 24, 36, and
    48 steps.
    }
    \label{fig:fusion_ablation}
\end{figure*}

\paragraph{Fusion and modality-specific regularization}
We next examine whether the multimodal trends depend on how visual and tactile features are fused and regularized.
Figure~\ref{fig:fusion_ablation} compares cross-attention, gated fusion, FiLM, and late concatenation, together with a late-concatenation variant that applies representation regularization only to the visual branch.
The fusion mechanisms achieve comparable performance in several image-based settings, while more complex fusion designs do not consistently improve the point-cloud-based models.
In contrast, late concatenation with visual-only regularization performs favorably across several modality combinations, including PointCloud+TacFF.

One possible explanation is that visual and tactile observations exhibit different temporal characteristics.
Visual information is available throughout the interaction, whereas tactile changes are more strongly concentrated around contact onset, release, and redistribution.
Applying the same regularization to both branches may therefore attenuate short-lived tactile variations that are useful for predicting contact evolution.
Based on these results, we use late concatenation with visual-only regularization as the default configuration.
This finding supports modality-specific regularization under the evaluated setting, rather than implying that tactile representations should remain unregularized in general.

\begin{figure*}[t]
    \centering
    \includegraphics[width=0.8\textwidth]{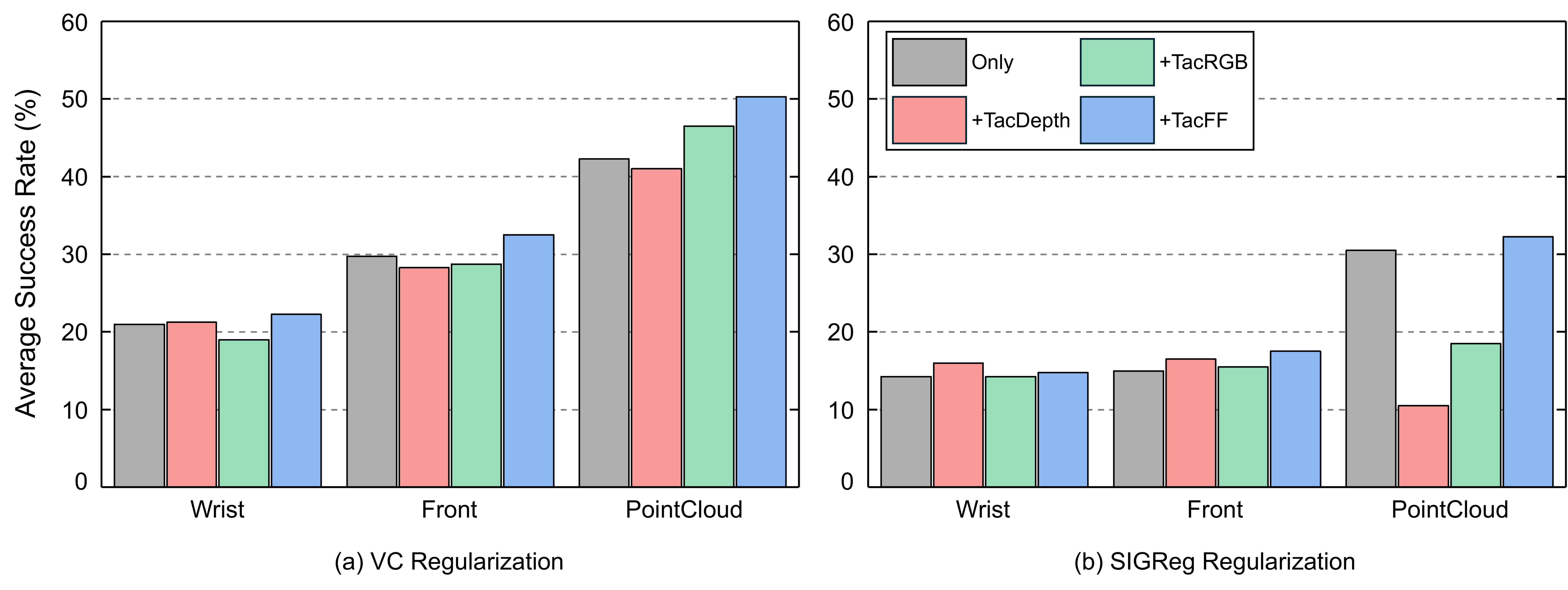}
    \caption{\textbf{Regularizer comparison on USB insertion.}
    Average planning success across 12-, 24-, 36-, and 48-step goal offsets for variance--covariance (VC) regularization and SIGReg across visual--tactile modality combinations.}
    \label{fig:reg_ablation}
\end{figure*}

\paragraph{Regularizer choice}
Finally, we compare VICReg-style variance--covariance (VC) regularization with
SIGReg~\cite{maes2026leworldmodel} across visual and tactile modality combinations on USB insertion (Fig.~\ref{fig:reg_ablation}).
Under the evaluated training and planning settings, VC produces higher average planning success than SIGReg across all modality combinations.
We therefore use VC in the main experiments.

\vspace{-2pt}
\section{Real-World Validation}
\label{sec:real_world}
We conduct 900 physical rollouts across six contact-rich tasks, two robot platforms, and three tactile sensors to evaluate whether the representation trends observed in simulation persist across physical settings.
On a Franka Panda equipped with GelSight Mini, we fix the visual input to PointCloud and compare vision-only, PointCloud+TacRGB, and PointCloud+TacFF across insertion, disassembly, screwing, and exploration tasks, providing a controlled evaluation of tactile representation.
On a reBot manipulator equipped with FlexiTac V2 and DM-Tac W2M, we evaluate combinations of wrist-view RGB, front-view RGB, and PointCloud with no tactile input, TacArray, or TacFF on peg insertion and occluded box search.
Together, these experiments test the identified representation trends across robot embodiments, gripper configurations, sensing hardware, and task observability conditions.

\begin{figure*}[t]
    \centering
    \includegraphics[width=1.0\textwidth]{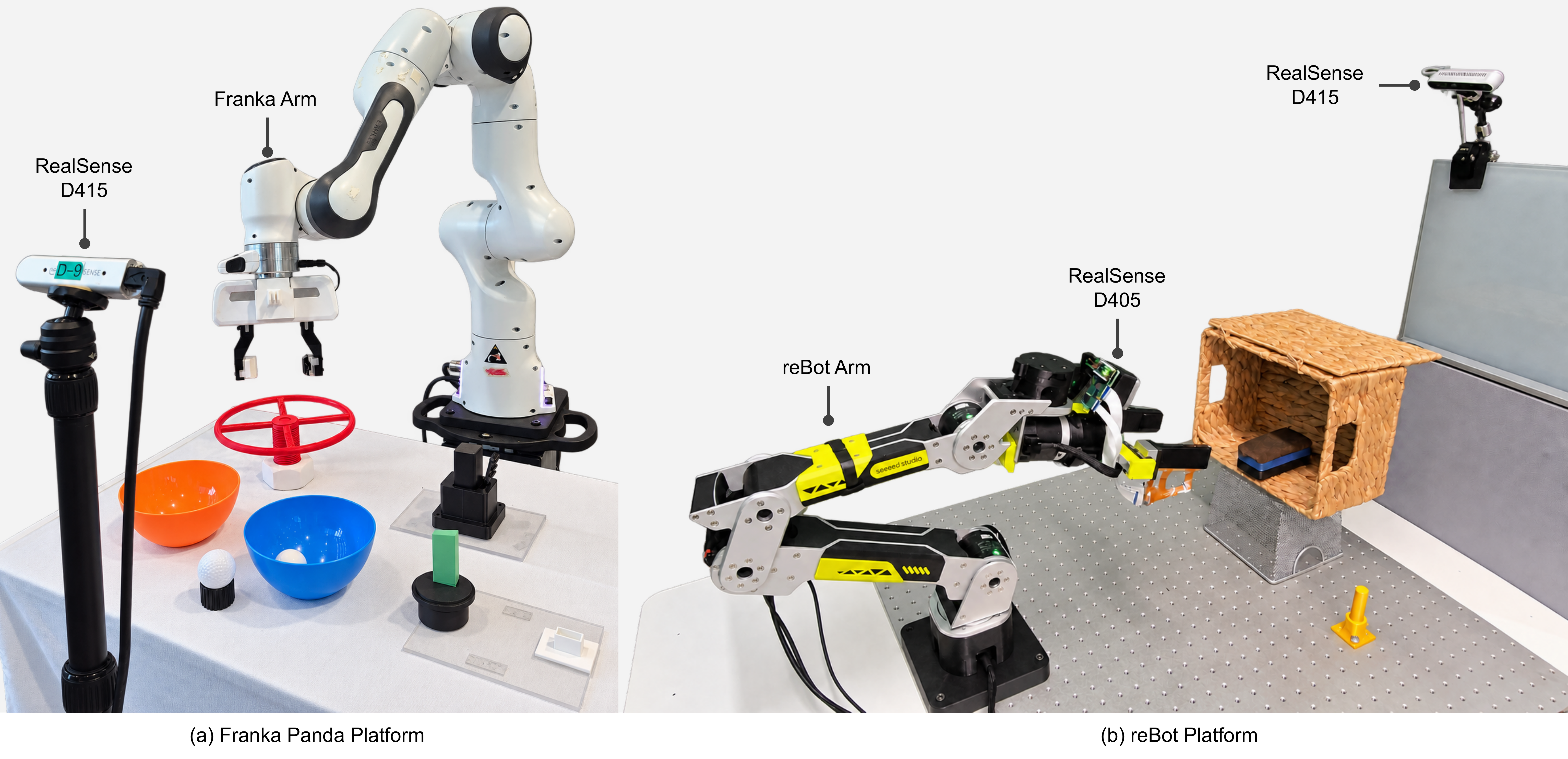}
    \caption{
    \textbf{Real-world experimental platforms.}
    (a) The Franka Panda platform uses a front-view Intel RealSense D415
    RGB-D camera and a parallel gripper equipped with tactile sensing.
    (b) The reBot platform uses a front-view RealSense D415 camera and a
    wrist-mounted RealSense D405 camera.
    }
    \label{fig:real_setup}
\end{figure*}

\subsection{Robot Platforms and Sensory Configurations}
\label{sec:real_setup}
We conduct real-world experiments on the two platforms shown in Fig.~\ref{fig:real_setup}, which differ in embodiment, control interface, camera configuration, gripper geometry, and tactile sensing hardware.

\begin{figure}[t]
    \centering
    \includegraphics[width=1.0\columnwidth]{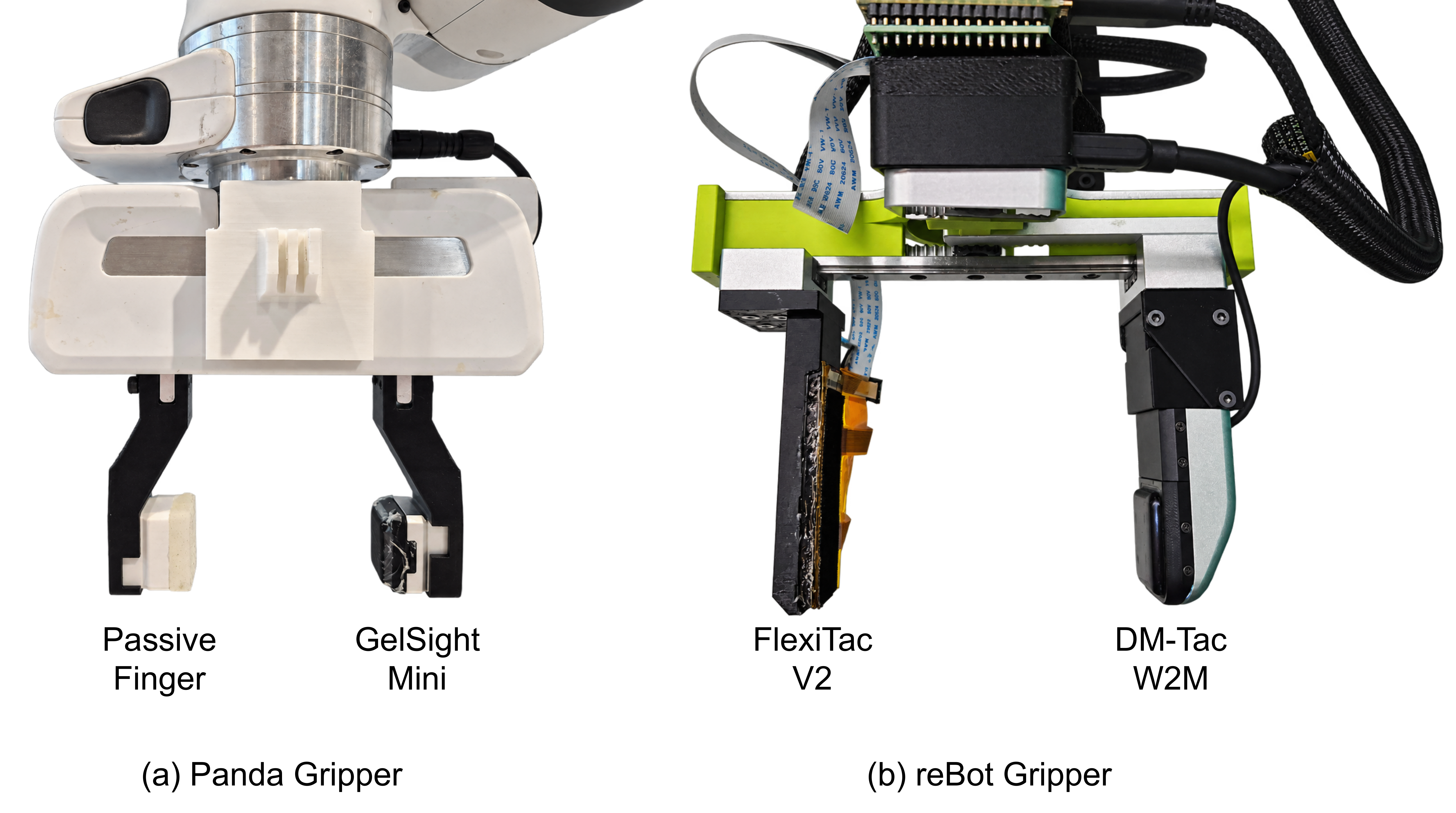}
\caption{\textbf{Real-world gripper and tactile configurations.}
(a) Panda gripper with a passive fingertip and GelSight Mini.
(b) reBot gripper with opposing FlexiTac V2 and DM-Tac W2M sensors.}
    \label{fig:gripper}
\end{figure}

\textbf{Franka Panda setup.}
The first setup uses a 7-DoF Franka Emika Panda~\cite{panda} controlled through Cartesian impedance control.
The robot is equipped with a front-view Intel RealSense D415 RGB-D camera~\cite{realsense} and a GelSight Mini tactile sensor~\cite{gelsightmini}.
Demonstrations are collected using a 6-DoF 3DConnexion SpaceMouse~\cite{spacemouse}, with observations and actions recorded at
10~Hz.

As shown in Fig.~\ref{fig:gripper}(a), the Panda gripper pairs one passive fingertip with a GelSight Mini.
GelSight Mini is a vision-based tactile sensor that captures surface appearance and sparse marker motion under contact.
We use its raw tactile images as TacRGB observations and construct TacFF by combining photometric surface reconstruction with tracked marker displacements.
Consequently, the Panda experiments compare TacRGB and TacFF derived from the same tactile hardware, allowing the effect of tactile representation to be evaluated while keeping the robot, sensor, visual input, and planning framework fixed.

\textbf{reBot setup.}
The second setup uses a reBot manipulator~\cite{reBot} controlled through position commands.
It is equipped with a wrist-mounted Intel RealSense D405 RGB-D camera, a front-view Intel RealSense D415 RGB-D camera, and a parallel gripper instrumented with two different tactile sensors.
As shown in Fig.~\ref{fig:gripper}(b), one finger carries a FlexiTac sensor~\cite{huang20243d}, while the other carries a DM-Tac W2M sensor~\cite{daimon_dmtac}.
Demonstrations are collected using a joystick at 10~Hz.

FlexiTac is a thin pressure-array sensor that directly measures a spatially organized contact-pressure field.
We use this pressure map as the TacArray representation, which provides a low-resolution description of local surface deformation.
DM-Tac W2M is a vision-based tactile sensor with a dense particle pattern, supporting dense marker tracking and contact-surface reconstruction.
We use its tracked contact motion to construct the TacFF representation.

In addition to the Panda experiments, we further evaluate the method on the reBot platform with a different kinematic configuration, where both the tactile sensor and its associated representation are changed.
It should therefore be interpreted as a cross-sensor evaluation of tactile sensor--representation configurations, rather than as a representation-only ablation.
Together, the two platforms test whether the observed trends remain consistent under changes in robot embodiment, control interface, gripper geometry, camera viewpoint, and tactile sensing hardware.

\textbf{Observation preprocessing.}
The wrist- and front-view RGB observations are acquired at $640\times480$ resolution.
Before encoding, each image is resized to $224\times224$, converted to floating point, and linearly mapped from $[0,255]$ to $[-1,1]$.

Point clouds are reconstructed from the front-view RGB-D observations using the calibrated camera intrinsics.
Because the raw point clouds contain substantial background geometry, we first apply workspace cropping and use RANSAC plane fitting~\cite{ransac} to remove the table surface.
The remaining geometry is downsampled to 1,024 points, each represented by its XYZ coordinates and RGB values.
The XYZ channels are standardized using the per-axis mean and standard deviation computed from the training episodes, while the RGB channels are mapped to $[-1,1]$.
This produces a $1024\times6$ PointCloud input consistent with that used in simulation.

GelSight Mini provides $240\times160$ tactile RGB observations.
Marker tracking produces a $7\times9$ displacement field, while the reconstructed surface deformation is sampled at the corresponding marker locations.
The resulting three-channel field, comprising depth and two in-plane motion components, is spatially interpolated to $10\times14\times3$ to match the TacFF resolution used in simulation.
FlexiTac produces a $12\times32$ pressure array, which is provided to the encoder as a single-channel tactile map.
For DM-Tac W2M, dense marker tracking and surface reconstruction are initially computed at $320\times240$ resolution and subsequently aggregated into the same $10\times14\times3$ TacFF format.
For both TacArray and TacFF, no-contact baselines and channel-wise scaling parameters are estimated from the training data before model training.

\subsection{Data Collection, Training, and Evaluation Protocol}
\label{sec:real_protocol}
For each of the six physical tasks, we collect 80 teleoperated demonstrations at 10~Hz.
The robot initial pose is varied across demonstrations, producing distinct initial observations and interaction trajectories.
The first 50 demonstrations are used for world-model training, while the remaining 30 are held out exclusively for physical evaluation.
Training and evaluation episodes are therefore disjoint.

All policies are trained for 200 epochs, and the final checkpoint is used for robot evaluation without task-specific checkpoint selection.
The encoder, predictor, representation regularization, multimodal fusion, and planning configurations follow the settings described in Secs.~\ref{sec:training_plannig} and \ref{sec:controlled_representation}.
Each held-out demonstration defines one evaluation condition.
We use its initial observation as the starting condition and select the observation 50 steps later as the goal.
The corresponding physical task configuration is recreated, after which the learned world model generates task-space actions under receding-horizon control.
All representation configurations are evaluated using the same 30 held-out initial--goal pairs for each task.

The Panda experiments evaluate three configurations: PointCloud, PointCloud+TacRGB, and PointCloud+TacFF.
This selection follows the simulation study: PointCloud is the strongest vision-only representation, while TacRGB and TacFF enable a controlled comparison of two representations derived from the same GelSight Mini sensor.
The evaluation covers one physical task from each of the four interaction categories.

The reBot experiments instead evaluate all combinations of three visual representations (wrist-view RGB, front-view RGB, and PointCloud) and three tactile settings (no tactile input, TacArray, and TacFF).
This $3\times3$ evaluation is conducted on both peg insertion and occluded box search, testing the representation trends under a different embodiment and tactile sensing configuration.
In total, the real-world study comprises 360 Panda rollouts and 540 reBot rollouts, for a total of 900 physical trials.

\subsection{Physical Tasks and Success Criteria}
\label{sec:real_tasks}
\begin{figure*}[t]
  \centering
  \includegraphics[width=1.0\linewidth]{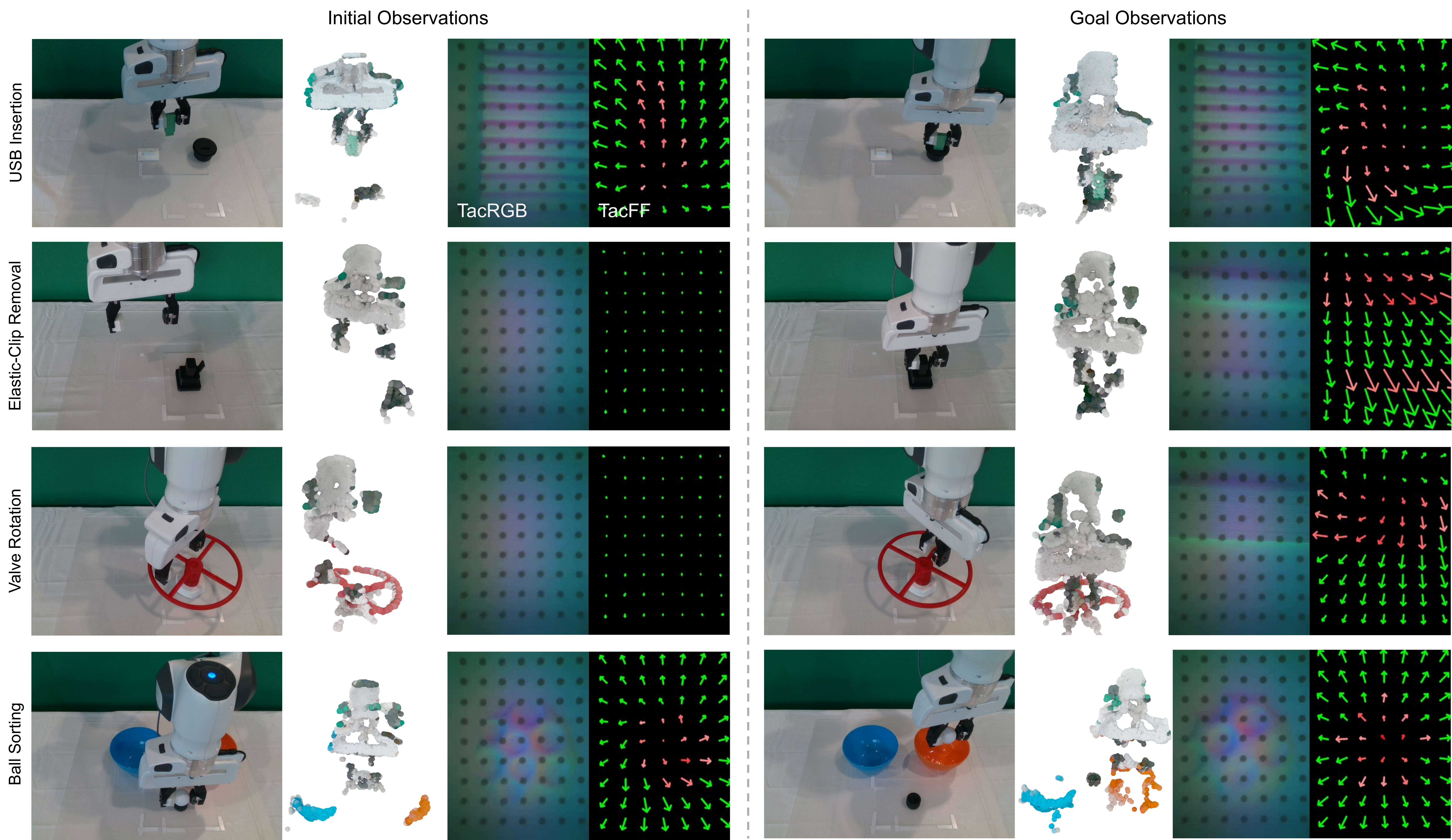}
  \caption{
  \textbf{Initial and goal multimodal observations for the real-world Panda tasks.}
  Rows show USB insertion, elastic-clip removal, valve rotation, and ball sorting, representing insertion, disassembly, screwing, and exploration, respectively.
  Within each initial and goal group, observations are ordered as front-view RGB, the corresponding RGB-D-derived point cloud, GelSight Mini TacRGB, and TacFF.
  Front-view RGB is included only to visualize the physical configuration; all Panda models use PointCloud as their visual input.
  }
  \label{fig:panda_tasks}
\end{figure*}

\begin{figure*}[t]
  \centering
  \includegraphics[width=1.0\linewidth]{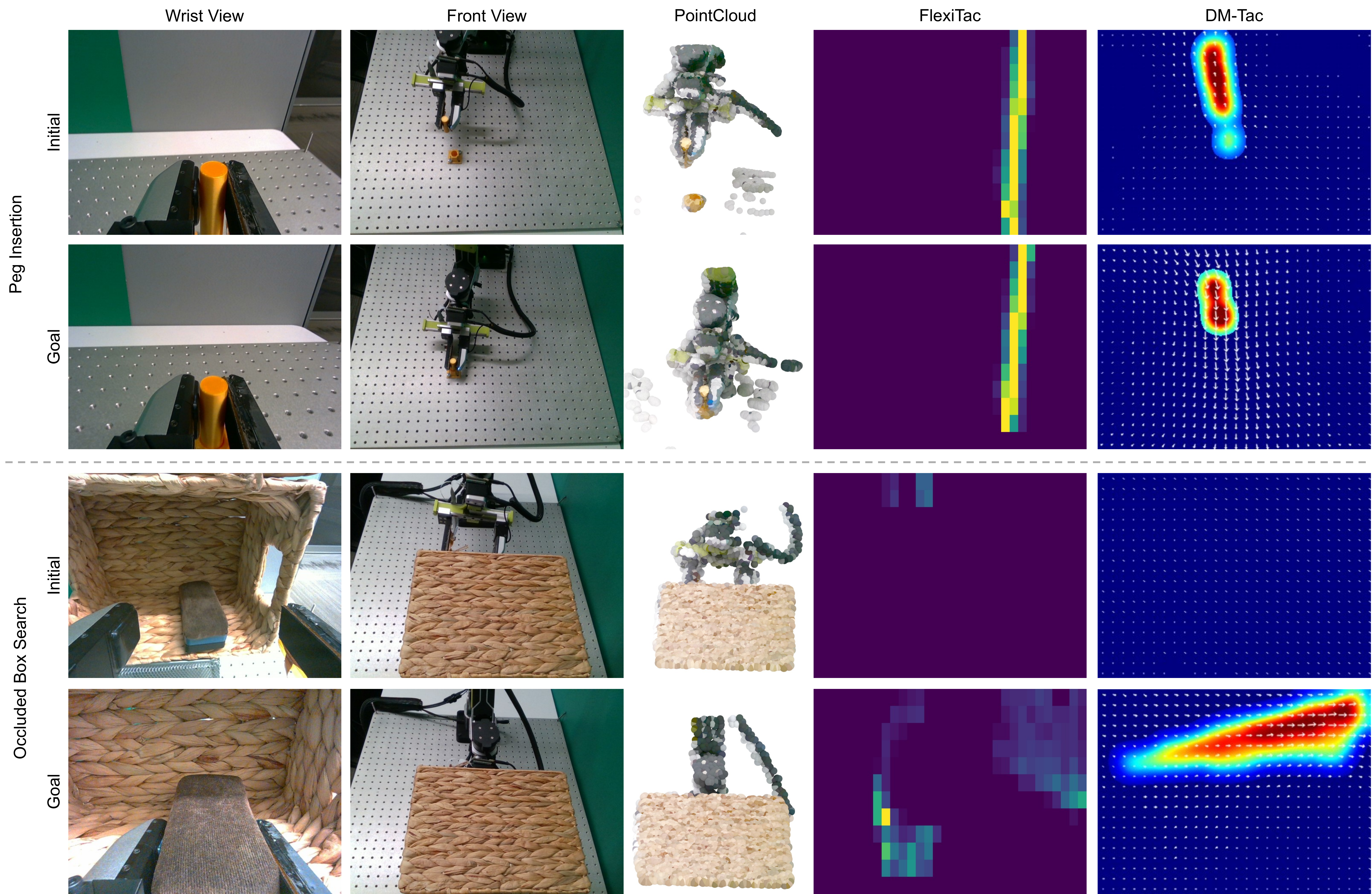}
  \caption{
  \textbf{Initial and goal multimodal observations for the real-world reBot tasks.}
  The top and bottom groups show peg insertion and occluded box search, respectively.
  Columns show wrist-view RGB, front-view RGB, the RGB-D-derived point cloud, the FlexiTac pressure field, and the structured deformation and shear field obtained from DM-Tac W2M.
  }
  \label{fig:reBot_tasks}
\end{figure*}

The Panda experiments cover the same four interaction categories as the simulation benchmark.
USB insertion represents constrained insertion; elastic-clip removal represents disassembly; valve rotation represents sustained rotational interaction; and ball sorting represents exploratory manipulation.
Representative initial and goal observations for these tasks are shown in Fig.~\ref{fig:panda_tasks}.
The reBot experiments include peg insertion, which requires contact-sensitive alignment and insertion, and occluded box search, in which the target object must be located and grasped inside a visually restricted workspace (Fig.~\ref{fig:reBot_tasks}).

For USB insertion, a trial is successful if the connector is fully seated in the target socket.
Elastic-clip removal is successful if the gripper securely grasps the clip and disengages it from the fixture.
Valve rotation is successful if the gripper maintains contact with the valve and rotates it to the prescribed goal orientation.
Ball sorting is successful if the robot identifies the grasped ball and places it into the corresponding target container.
Peg insertion is successful if the peg is fully inserted into the target hole.
Occluded box search is considered successful if the robot locates the eraser and establishes a stable grasp, which is verified by an automatic lift executed after the grasp.

\subsection{Controlled Panda Evaluation}
\label{sec:panda_results}
On the Panda platform, we fix the visual representation to PointCloud and compare three configurations: PointCloud-only, PointCloud+TacRGB, and PointCloud+TacFF.
Because TacRGB and TacFF are derived from the same GelSight Mini sensor, this setup separates the effect of tactile representation from changes in the robot, tactile hardware, visual observation, world model, and planning procedure.

\begin{table}[t]
\centering
\caption{
\textbf{Controlled real-world representation comparison on the Panda platform.}
}
\label{tab:panda_results}
\setlength{\tabcolsep}{7pt}
\renewcommand{\arraystretch}{1.18}
\begin{tabular}{lccc}
\toprule
\textbf{Task}
& \textbf{PC}
& \textbf{PC+TacRGB}
& \textbf{PC+TacFF} \\
\midrule
USB Insertion & 2/30 & 6/30 & \textbf{9/30} \\
Elastic-Clip Removal & 9/30 & 15/30 & \textbf{18/30} \\
Valve Rotation & 11/30 & 14/30 & \textbf{19/30} \\
Ball Sorting & 16/30 & 19/30 & \textbf{21/30} \\
\midrule
\textbf{Average (\%)} & 31.7 & 45.0 & \textbf{55.8} \\
\bottomrule
\end{tabular}

\vspace{3pt}
\parbox{0.9\linewidth}{
\footnotesize
Each task--configuration pair is evaluated over 30 physical trials.
All configurations use the same held-out initial--goal pairs.
}
\end{table}

As shown in Table~\ref{tab:panda_results}, both tactile configurations improve the average success rate over PointCloud-only.
PointCloud achieves an average success rate of 31.7\%, which increases to 45.0\% with TacRGB and 55.8\% with TacFF.
The corresponding improvements over the vision-only baseline are 13.3 and 24.1 percentage points, respectively.
PointCloud+TacFF also achieves the highest success rate on each of the four physical tasks, outperforming PointCloud+TacRGB by 10.8 percentage points on average.

The magnitude of the tactile benefit varies across interaction categories.
Relative to PointCloud-only, PointCloud+TacFF improves USB insertion from 2/30 to 9/30, elastic-clip removal from 9/30 to 18/30, valve rotation from 11/30 to 19/30, and ball sorting from 16/30 to 21/30.
The largest absolute improvements occur in elastic-clip removal and valve rotation, whereas ball sorting has a stronger vision-only baseline and shows a smaller, but still positive, tactile gain.

These physical experiments support the principal trend observed in simulation: point-cloud-based predictive planning benefits from tactile observations, and the magnitude of this benefit depends on how contact information is represented.
Under the controlled Panda setting, the structured TacFF representation provides larger and more consistent gains than TacRGB across all four interaction categories.

\subsection{Cross-Platform and Cross-Sensor Validation}
\label{sec:rebot_results}
We next evaluate whether the observed representation trends persist under a different robot embodiment, gripper geometry, and tactile sensing system.
Unlike the Panda experiments, which fix the visual representation to PointCloud, the reBot experiments evaluate the complete $3\times3$ combination of visual and tactile representations.

\begin{table*}[t]
\centering
\caption{
\textbf{Cross-representation and cross-sensor evaluation on the reBot platform.}
}
\label{tab:rebot_results}
\setlength{\tabcolsep}{8pt}
\renewcommand{\arraystretch}{1.15}
\begin{tabular}{lccc|ccc}
\toprule
& \multicolumn{3}{c|}{\textbf{Peg Insertion}}
& \multicolumn{3}{c}{\textbf{Occluded Box Search}} \\
\cmidrule(lr){2-4}
\cmidrule(lr){5-7}
\textbf{Visual Representation}
& Only & +TacArray & +TacFF
& Only & +TacArray & +TacFF \\
\midrule
Wrist View
& 8/30  & 9/30  & 13/30
& 9/30  & 11/30 & \textbf{15/30} \\
Front View
& 13/30 & 22/30 & 21/30
& 1/30  & 1/30  & 5/30 \\
PointCloud
& 18/30 & 17/30 & \textbf{24/30}
& 2/30  & 2/30  & 3/30 \\
\bottomrule
\end{tabular}

\vspace{3pt}
\parbox{0.65\textwidth}{
\footnotesize
Each entry reports successful trials out of 30 runs. The best result for each task is shown in bold.
All configurations use the same 30 held-out initial--goal pairs. TacArray is obtained from FlexiTac V2, while TacFF is obtained from DM-Tac W2M.
}
\end{table*}

As shown in Table~\ref{tab:rebot_results}, the relative effectiveness of the visual representations depends strongly on the task.
For peg insertion, PointCloud provides the strongest vision-only performance at 18/30 and reaches the highest overall success rate of 24/30 when combined with TacFF.
Front-view RGB also benefits substantially from tactile sensing, improving from 13/30 to 22/30 with TacArray and 21/30 with TacFF.
In contrast, TacArray does not improve the PointCloud baseline, demonstrating that adding tactile input is not uniformly beneficial.

Occluded box search produces a different visual ranking.
Because the interaction occurs inside an occluded workspace, wrist-view RGB provides the strongest vision-only baseline at 9/30, compared with 1/30 for front-view RGB and 2/30 for PointCloud.
Combining the wrist view with TacFF increases success to 15/30, the highest result on this task.
TacFF also improves the front-view model from 1/30 to 5/30, although the absolute success rate remains limited by the weak visibility of the external view.

These results extend the simulation and Panda findings to a different robot and tactile sensing setup, while also showing that the strongest visual representation depends on the observability structure of the task.

\begin{figure*}[t]
  \centering
  \includegraphics[width=1.0\linewidth]{figures/analysis.pdf}
  \caption{
  \textbf{Representative real-world rollouts with and without tactile sensing.}
TacFF enables correction of residual misalignment during USB insertion and
stabilization of the grasp during occluded box search, succeeding where the
corresponding vision-only models fail.
RGB frames are shown for clarity; model inputs are specified by the row labels,
and insets show TacFF responses during contact.}
  \label{fig:analysis}
\end{figure*}

Figure~\ref{fig:analysis} illustrates how these aggregate trends appear within individual interactions.
In USB insertion, the PointCloud-only model reaches the socket but fails to recover when residual misalignment produces contact at the opening.
With TacFF, the evolving deformation and shear pattern indicates the direction of contact imbalance, enabling corrective actions that complete the insertion.
In occluded box search, the wrist view maintains local visibility inside the enclosure but does not reliably indicate whether the object is stably constrained between the fingers.
TacFF helps distinguish transient contact from a stable grasp, supporting correction before lifting.

Although individual examples do not establish causality, they are consistent with the task-level results: explicit geometry is effective when the relevant scene structure is observable, whereas tactile feedback becomes especially valuable when contact states are visually ambiguous.
Together, the Panda and reBot experiments show that these representation effects persist across robot embodiments and tactile sensing systems, without implying that a single modality combination is universally optimal.

\section{Conclusion and Limitations}

\paragraph{Conclusion}
We presented \textbf{ContactWorld}, a systematic investigation of sensory representations for vision--tactile world models in contact-rich manipulation.
Our results show that predictive planning depends not only on the information available to the model, but also on whether its representation preserves interaction geometry, reflects physical motion, and supports stable multistep prediction.
Point clouds provide a strong representation when task-relevant geometry is observable, while tactile sensing is most useful when it exposes contact information that is ambiguous or absent in the visual input.
Accordingly, tactile observations do not provide uniform gains across visual modalities or tasks: structured TacFF yields the most consistent tactile benefits, whereas the strongest visual--tactile combination depends on task observability.

These conclusions are supported by representation diagnostics, controlled ablations, and 900 physical trials across six tasks, two robot platforms, and three tactile sensing systems.
Together, the results identify representation design and cross-modal complementarity as central factors in predictive control, and provide empirical guidance for constructing vision--tactile world models for contact-rich manipulation.

\paragraph{Limitations}
ContactWorld primarily considers individual manipulation skills over moderate temporal horizons.
Although appropriate sensory representations slow predictive degradation, performance still decreases as prediction and planning extend farther into the future.
Whether the observed representation trends persist in compositional tasks requiring repeated contact transitions, recovery, and higher-level decision making therefore remains an open question.

The physical experiments span multiple platforms and tactile sensing systems, but do not fully disentangle representation from sensor mechanics.
In particular, TacArray and TacFF on the reBot are obtained from different sensors, and separate models are trained for each hardware configuration.
Evaluating multiple representations derived from the same tactile hardware, together with transfer across sensors and robot embodiments, will be necessary to determine which representation principles generalize across sensing systems.

\bibliographystyle{IEEEtran}
\bibliography{ContactWorld}

\end{document}